\documentclass{article}
\PassOptionsToPackage{numbers, sort&compress}{natbib}
\usepackage[final]{neurips_2022}

\usepackage[utf8]{inputenc} %
\usepackage[T1]{fontenc} %
\usepackage{url} %
\usepackage{booktabs} %
\usepackage{amsfonts} %
\usepackage{nicefrac} %
\usepackage{microtype} %
\usepackage[dvipsnames]{xcolor}
\usepackage{tikz}
\usepackage[framemethod=tikz]{mdframed}

\usepackage{microtype}
\usepackage{graphicx}
\usepackage{subfigure}
\usepackage{booktabs} %

\usepackage{amsmath}
\usepackage{amssymb}
\usepackage{amsthm}
\usepackage{xcolor}
\usepackage{tabularx}
\usepackage{makecell}
\newtheorem{theorem}{Theorem}

\usepackage{url} %
\usepackage{booktabs} %
\usepackage{amsfonts} %
\usepackage{nicefrac} %
\usepackage{microtype} %
\usepackage{multirow} %
\usepackage{algorithm}
\usepackage{algorithmic}

\usepackage{wrapfig}
\usepackage{comment}
\usepackage{amsmath,amssymb} %
\usepackage{symbols}
\usepackage{multirow}
\usepackage{ctable} %
\usepackage{pifont} %
\newcommand{\cmark}{\ding{51}}
\newcommand{\xmark}{\ding{55}}

\usepackage{amsmath,amsfonts,bm}

\def\floor#1{\lfloor #1 \rfloor}
\def\1{\bm{1}}

\def\sp{space}

\def\rvk{{\mathbf{k}}}

\def\vb{{\bm{b}}}

\def\vk{{\bm{k}}}

\def\vu{{\bm{u}}}

\def\vx{{\bm{x}}}
\def\vy{{\bm{y}}}
\def\vz{{\bm{z}}}

\def\mI{{\bm{I}}}

\def\mT{{\bm{T}}}

\DeclareMathAlphabet{\mathsfit}{\encodingdefault}{\sfdefault}{m}{sl}
\SetMathAlphabet{\mathsfit}{bold}{\encodingdefault}{\sfdefault}{bx}{n}

\def\sR{{\mathbb{R}}}

\def\sZ{{\mathbb{Z}}}

\DeclareMathOperator*{\argmax}{arg\,max}

\definecolor{darkred}{rgb}{0.7,0.1,0.1}
\definecolor{darkgreen}{rgb}{0.1,0.7,0.1}
\definecolor{cyan}{rgb}{0.7,0.0,0.7}
\definecolor{dblue}{rgb}{0.2,0.2,0.8}
\definecolor{maroon}{rgb}{0.76,.13,.28}
\definecolor{burntorange}{rgb}{0.81,.33,0}
\definecolor{tealblue}{rgb}{0.212,0.459, 0.533}

\definecolor{mypink}{rgb}{0.93359375, 0.62109375, 0.83984375}

\definecolor{pp}{rgb}{0.43921569, 0.18823529, 0.62745098}
\definecolor{rr}{rgb}{0.5254902 , 0.00784314, 0.12941176}
\definecolor{bb}{rgb}{0.09019608, 0.23529412, 0.37647059}
\definecolor{yy}{rgb}{0.49803922, 0.3372549 , 0.0}
\definecolor{gg}{rgb}{0.02352941, 0.3372549 , 0.17647059}

\newcommand{\poly}[1]{\text{Poly}(#1)}

\usepackage[multiple]{footmisc}
\usepackage[shortlabels]{enumitem}
\usepackage{wrapfig}

\definecolor{mybrown}{rgb}{0.87058824, 0.56078431, 0.01960784}
\definecolor{myblue}{rgb}{0.3372549 , 0.70588235, 0.91372549}
\definecolor{mypurple}{rgb}{0.8, 0.47058824, 0.7372549 }
\definecolor{myorange}{rgb}{0.835, 0.368, 0}
\definecolor{mygreen}{rgb}{0.00784314, 0.61960784, 0.45098039}
\definecolor{mygt}{rgb}{0.0078125 , 0.57421875, 0.40625}
\definecolor{mysp}{rgb}{0.84765625, 0.515625  , 0.0234375}
\definecolor{mycitecolor}{rgb}{0,0.08,0.45}
\definecolor{mygr}{rgb}{0.9607,0.9607,0.9607}
\definecolor{myoo}{rgb}{0.992,0.9176,0.9019}

\definecolor{myrr}{HTML}{AE031A}
\definecolor{mybb}{HTML}{0155B3}

\usepackage{pifont}
\usepackage[T1]{fontenc}
\usepackage[utf8]{inputenc}
\usepackage{pmboxdraw}
\usepackage[font={footnotesize}, labelfont=bf,labelsep=period]{caption}
\usepackage{thm-restate}
\usepackage[pagebackref=true,breaklinks=true,colorlinks, citecolor=mybb]{hyperref}

\mdfdefinestyle{MyFrame}{
    linecolor=Black,
    outerlinewidth=0.1pt,
    roundcorner=2pt,
    innertopmargin=0.4\baselineskip,
    innerbottommargin=0.4\baselineskip,
    innermargin = 0.2cm,
    outermargin = 0.cm,
    backgroundcolor=Gray!0!white}
\usepackage[multiple]{footmisc}
\newtheorem{lemma}[theorem]{Lemma}

\usepackage[textsize=tiny]{todonotes}

\usepackage[frozencache,cachedir=minted-cache]{minted}
\usepackage{listings}

\newcommand{\papertitle}{Learnable Polyphase Sampling for Shift Invariant and Equivariant Convolutional Networks}
\title{\papertitle}

\author{
  Renan A. Rojas-Gomez\thanks{Equal contribution.} \quad\quad Teck-Yian Lim\footnotemark[1] \\
  \textbf{Alexander G. Schwing} \quad\quad  \textbf{Minh N. Do} \quad\quad \textbf{Raymond A. Yeh}\textsuperscript{$\dagger$}\\ \\
  Department of Electrical Engineering, University of Illinois at Urbana-Champaign\\
  \textsuperscript{$\dagger$}Department of Computer Science, Purdue University
}

\begin{document}
\maketitle
\setcounter{footnote}{0}
\begin{abstract}
We propose learnable polyphase sampling (LPS), a pair of learnable down/upsampling layers that enable truly shift-invariant and equivariant convolutional networks. LPS can be trained end-to-end from data and generalizes existing handcrafted downsampling layers. It is widely applicable as it can be integrated into any convolutional network by replacing down/upsampling layers. We evaluate LPS on image classification and semantic segmentation. Experiments show that LPS is on-par with or outperforms existing methods in both performance and shift consistency. For the first time, we achieve true shift-equivariance on semantic segmentation (PASCAL VOC), \textit{i.e.}, 100\% shift consistency, outperforming baselines by an absolute 3.3\%. Our project page and code are available at \url{https://raymondyeh07.github.io/learnable_polyphase_sampling/}

\end{abstract}

\section{Introduction}\label{sec:intro}
For tasks like image classification, shifts of an object do not change the corresponding object label, \ie, the task is shift-invariant. This shift-invariance property has been incorporated into deep-nets yielding  convolutional neural nets (CNN). Seminal works on CNNs~\cite{Fukushima_1980, lecun1999object} directly attribute the model design to shift-invariance. For example,~\citet{Fukushima_1980} states ``the network has an ability of position-invariant pattern recognition'' and~\citet{lecun1999object} motivate CNNs by stating that they ``ensure some degree of shift invariance.'' 

CNNs have evolved since their conception. Modern deep-nets contain more layers, use different non-linearities and pooling layers. Re-examining these modern architectures, \citet{zhang2019making} surprisingly finds that modern deep-nets are not shift-invariant. To address this,~\citet{zhang2019making} and~\citet{zou2020delving} propose to perform anti-aliasing before each downsampling layer, and found it to improve the degree of invariance. More recently,~\citet{chaman2021truly} show that deep-nets can be ``truly shift-invariant,'' \ie, a model's output is identical for given shifted inputs. For this, they replace all downsampling layers with their adaptive polyphase sampling (APS) layer. 

While APS achieves true shift-invariance by selecting the max-norm polyphase component (a {handcrafted} downsampling scheme), an important question arises: are there more effective downsampling schemes that can achieve true shift-invariance? Consider an extreme case, a {handcrafted} deep-net that always outputs zeros is truly shift-invariant, but does not accomplish any task. This motivates to study how truly shift-invariant downsampling schemes can be {learned from data}.

For this we propose Learnable Polyphase Sampling (LPS), a pair of down/upsampling layers that yield truly shift-invariant/equivariant deep-nets and can be trained in an end-to-end manner. 
For downsampling, LPS can be easily integrated into existing deep-net architectures by swapping out the pooling/striding layers. Theoretically, LPS generalizes APS to downsampling schemes that cannot be represented by APS. Hence, LPS's ideal performance is never worse than that of APS. For upsampling, LPS guarantees architectures that are truly shift-equivariant, \ie, the output shifts accordingly when the input shifts. This is desirable for tasks like semantic image segmentation. 

To validate the proposed LPS, we conduct extensive experiments: (a) image classification on CIFAR10~\cite{krizhevsky2009learning} and ImageNet~\cite{deng2009imagenet}; (b) semantic segmentation on PASCAL VOC~\cite{everingham2015pascal}. We observe that the proposed approach outperforms APS and further improves anti-aliasing methods on both model performance and shift consistency.

{\bf \noindent Our contributions are as follows:}
\vspace{-0.1cm}
\begin{itemize}[leftmargin=0.45cm]
\vspace{-0.08cm}
\itemsep0em 
\item We propose learnable polyphase sampling (LPS), a pair of novel down/upsampling layers, and prove that they yield truly shift-invariant/-equivariant deep-nets. Different from prior works, our sampling scheme is trained from end-to-end and not handcrafted.
\item  We theoretically prove that LPS (downsampling) is a generalization of APS. Hence, in theory, LPS improves upon APS.
\item We conduct extensive experiments demonstrating the effectiveness of LPS on image classification and segmentation over three datasets comparing to APS and anti-aliasing approaches.
\end{itemize}
\vspace{-0.05cm}

\begin{figure*}[t]
\centering
\includegraphics[width=0.98\textwidth,
trim={0 0 0 4cm},clip
]{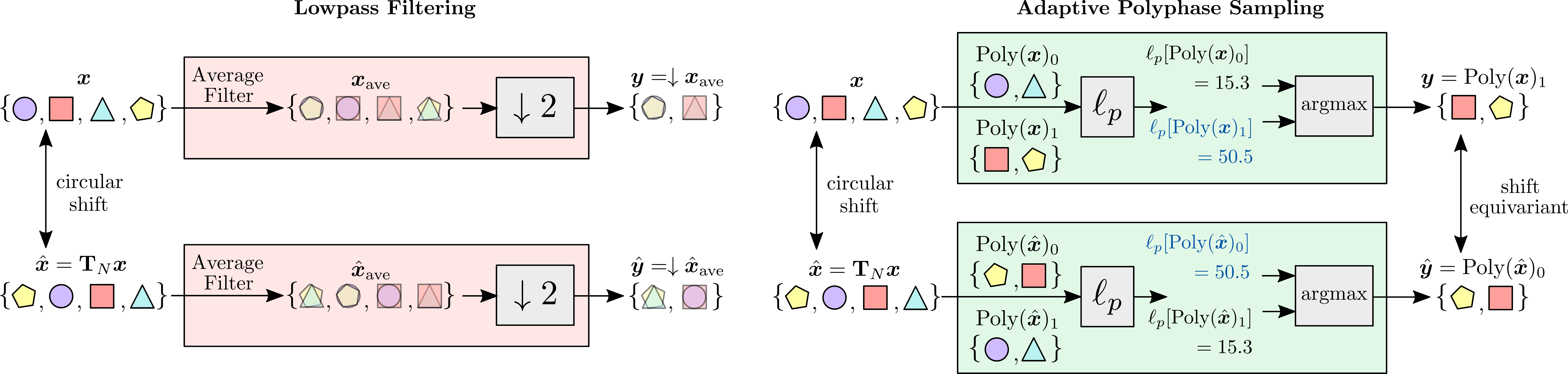}
\vspace{-.3\baselineskip}
\small
\caption{{\bf Left: lowpass filtering (LPF)}~\citep{zhang2019making} averages the inputs before downsampling to encourage shift-equivariance. The outputs $\vy$ and $\hat{\vy}$ are ``more similar'' due to the average filter. Despite improving shift consistency, the model is still shift variant by definition. {\bf Right:  adaptive polyphase sampling (APS)}~\citep{chaman2021truly} selects the component with the max $\ell_p$-norm. This hand-crafted rule is shift equivariant as it selects the square and pentagon in both $\hat\vy$ and $\vy$.}
\vspace{-0.5cm}
\label{fig:related}
\end{figure*}

\section{Related Work}\label{sec:rel}
In this section, we briefly discuss  related work, including shift invariant/equivariant deep-nets and pooling layers. Additional necessary concepts are reviewed in~\secref{sec:prelim}.

{\noindent \bf Shift Invariant/equivariant convolutional networks.}
Modern convolutional networks use striding or pooling to reduce the amount of memory and computation in the model~\cite{Krizhevsky_Sutskever_Hinton_2012,Simonyan_Zisserman_2015,He_Zhang_Ren_Sun_2016,Sandler_Howard_Zhu_Zhmoginov_Chen_2018}. As pointed out by~\citet{Azulay_Weiss_2019} and~\citet{zhang2019making}, these pooling/striding layers break the shift-invariance property of deep-nets. To address this issue,~\citet{zhang2019making} proposed to perform anti-aliasing, \ie, lowpass filtering (LPF) before each downsampling, a canonical signal processing technique for multi-rate systems~\cite{vetterli2014foundations}. We illustrate this approach in~\figref{fig:related} (left).~\citet{zou2020delving} further improved the LPF technique by using adaptive filters which better preserve edge information.

While anti-aliasing filters are effective,~\citet{chaman2021truly} show that true shift-invariance, \ie, 100\% shift consistency, can be achieved without anti-aliasing. Specifically, they propose Adaptive Polyphase Sampling (APS) which selects the downsampling indices, \ie, polyphase components, based on the $\ell_p$-norm of the polyphase components; a handcrafted rule, as illustrated in~\figref{fig:related} (right). In a follow up technical report~\cite{Chaman_Dokmanic_2021}, APS is extended to upsampling using unpooling layers~\cite{Zeiler_Fergus_2014, Badrinarayanan_Kendall_Cipolla_2017}, where the downsampling indices are saved to place values back to their corresponding spatial location during upsampling. Our work presents a novel pair of shift-invariant/equivariant down/upsampling layers which are trainable, in contrast to APS's handcrafted selection rule.

We note that generalizations of equivariance beyond shifts have also been studied~\cite{cohen2016group,bronstein2017geometric,ravanbakhsh2017equivariance,weiler2019general,venkataraman2019building,romero2020attentive,yeh2022equivariance,shakerinava21a} and applied to various domains, \eg, sets~\cite{ravanbakhsh_sets,zaheer2017deep, qi2017pointnet, hartford2018deep,yeh2019chirality,maron2020learning}, graphs~\cite{shuman2013emerging,defferrard2016convolutional,kipf2017semi,maron2018invariant,yeh2019diverse,dehaan2020gauge,liu2020pic,liu2021semantic,morris22a}, spherical images~\cite{cohen2018spherical,kondor2018clebsch,cohen2019gauge}, volumetric data~\cite{weiler20183d},~\etc. In this work, we focus solely on shift-equivariance for images with CNNs.

{\noindent \bf Pooling layers.} Many designs for better downsampling or pooling layers have been proposed. Popular choices are Average-Pooling~\cite{LeCun_Boser_Denker_Henderson_Howard_Hubbard_Jackel_1989} and Max-Pooling~\cite{Ranzato_Boureau_LeCun_others_2007}. Other generalizations also exists, \eg, $L_P$-Pooling~\cite{Sermanet_Chintala_LeCun_2012} which generalizes pooling to use different norms. The effectiveness of different pooling layers has also been studied by~\citet{Scherer_Muller_Behnke_2010}. More similar to our work is Stochastic-Pooling~\cite{zeiler2013stochastic} and Mixed Max-Average Pooling~\cite{Lee_Gallagher_Tu_2016}. Stochastic-Pooling constructs a probability distribution by normalizing activations within a window and sampling during training. In our work, we present a novel design which learns the sampling distribution. Mixed Max-Average Pooling learns a single scalar to permit a soft-choice between  Max- and Average-Pooling. In contrast, our  LPS has shift-equivariance guarantees while being end-to-end trainable.

\section{Preliminaries}\label{sec:prelim}
We provide a brief review on equivariant and invariant functions to establish the notation. For readability, we use one-dimensional data to illustrate these ideas. In practice, these concepts are generalized to multiple channels and two-dimensional data.

{\noindent\bf Shift invariance and equivariance.} 
The concept of \textit{equivariance}, a generalization of invariance, describes how a function's output is transformed given that the input is transformed in a \textit{predefined way.} For example, shift equivariance describes how the output is shifted given that the input is also shifted: think of image segmentation, if an object in the image is shifted then its corresponding mask is also shifted. 

A function $f: \sR^{N} \mapsto \sR^{M}$ is \textit{$\mT_N,\{\mT_M, \mI\}$-equivariant} (shift-equivariant) if and only if {\it (iff)}
\be\label{eq:shift-eq-def}
\hspace{-0.25cm}\exists~T \in \{\mT_M, \mI\} \text{~~s.t.~~} f(\mT_N\vx) = T f(\vx)~\forall \vx \in \sR^{N},
\ee
where $\mT_N\vx[n] \triangleq \vx[(n+1) {\;\tt mod\;} N]\ \forall n \in \sZ$ denotes a circular shift, $[\cdot]$ denotes the indexing operator, and $\mI$ denotes the identity function. This definition of equivariance handles the ambiguity that arises when shifting by one and downsampling by two. Ideally, a shift by one at the input should result in a 0.5 shift in the downsampled signal, which is not achievable on the integer grid. Hence, this definition considers \textit{either a shift by one or a no shift} at the output as equivariant. 

Following the equivariance definition, invariance can be viewed as a special case where the transformation at the output is an identity function, $\mI$. Concretely, a function $f: \sR^{N} \mapsto \sR^{M}$ is $\mT_N,\{\mI\}$-equivariant (shift-invariant) {\it iff}
\be
f(\mT_N\vx) = f(\vx) \;\forall\vx \in \sR^{N}.
\ee

To obtain shift-invariance from shift-equivariant functions it is common to use global pooling. Observe that 
\be\sum_{m} f(\mT\vx)[m] = \sum_{m} (\mT f(\vx))[m]
\ee is shift-invariant if $f$ is shift-equivariant, as summation is an orderless operation. Note that the composition of shift-equivariant functions maintains shift-equivariance. Hence, $f$ can be a stack of equivariant layers, \eg, a composition of convolution layers.

While existing deep-nets~\cite{lin2013network, Sandler_Howard_Zhu_Zhmoginov_Chen_2018, He_Zhang_Ren_Sun_2016} do use global spatial pooling, these architectures are \textit{not} shift-invariant. This is due to pooling and downsampling layers, which are not shift-equivariant as we review next. 

{\noindent\bf Downsampling and pooling layers.}
A downsampling-by-two layer $\text{D}: \sR^{N}  \mapsto \sR^{\floor{N/2}}$ is defined as
\bea
\text{D}(\vx)[n] = \vx[2n] \; \forall n \in \sZ,
\eea
which returns the even indices of the input $\vx$. As a shift operator makes the odd indices even, a downsampling layer is not shift-equivariant/invariant. 

Commonly used average or max pooling can be viewed as an average or max filter followed by downsampling, hence pooling is also not shift-equivariant/invariant. To address this issue,~\citet{chaman2021truly} propose adaptive polyphase sampling (APS) which is an input dependent (adaptive) selection of the odd/even indices.

{\noindent\bf Adaptive polyphase sampling.}
Proposed by~\citet{chaman2021truly}, adaptive polyphase sampling (APS) returns whether the odd or even indices, \ie, the polyphase components, based on their norms. Formally, $\text{APS}: \sR^{N} \mapsto \sR^{\floor{N/2}}$ is defined as:
\bea
\text{APS}(\vx) = \begin{cases}
      \poly{\vx}_0 & \text{if } \norm{\poly{\vx}_0} > \norm{\poly{\vx}_1}\\
      \poly{\vx}_1 & \text{otherwise}
    \end{cases},
\eea
where $\vx \in \sR^{N}$ is the input and $\poly{\vx}_{i}$ denotes the polyphase components, \ie, 
\bea
\poly{\vx}_0[n] = \vx[2n]~~\text{and}~~\poly{\vx}_1[n] = \vx[2n+1].
\eea
While this handcrafted selection rule achieves a consistent selection of the polyphase components, it is {\it not the only way} to achieve it, \eg, returning the polyphase component with the smaller norm. In this work, we study a family of shift-equivariant sampling layers and propose how to learn them in a data-driven manner.

\begin{figure*}
\centering
\includegraphics[width=0.97\textwidth,
trim={0 0 0 10cm},clip
]{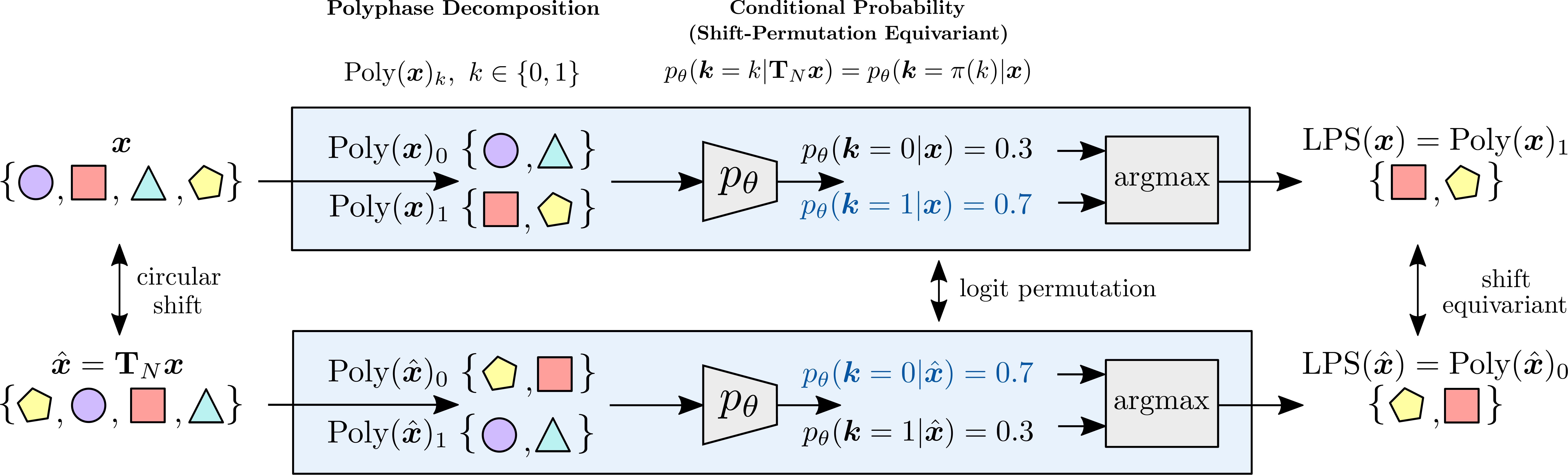}
\small
\caption{Illustration of our proposed LPD design. 
Inspired by the polyphase permutation property, \textit{Lemma 1}, the likelihood of selecting a polyphase component is learned with a shift-permutation equivariant CNN model $p_{\theta}$. This enforces components between the input and its shifted version to have consistent logits, which leads to a shift-equivariant downsampling layer.} 
\vspace{-0.4cm}
\label{fig:pipeline}
\end{figure*}

\section{Approach}\label{sec:app}
Our goal is to design a learnable down/upsampling layer that is shift-invariant/equivariant. We formulate down/upsampling by modeling the conditional probability of selecting each polyphase component given an input. For this we use a small neural network. This enables the sampling scheme to be trained end-to-end from data, hence the name learnable polyphase sampling (LPS). 

In~\secref{subsec:lps}, we introduce learnable polyphase downsampling (LPD), discuss how to train it end-to-end, and show that it generalizes APS. In~\secref{subsec:lps_design}, we propose a practical layer design of LPD. Lastly, in~\secref{subsec:extension}, we discuss how to perform LPS for upsampling, namely, learnable polyphase upsampling (LPU). For readability, we present the approach using one dimensional data, \ie, a row in an image.

\subsection{Learnable Polyphase Downsampling}\label{subsec:lps}
We propose learnable polyphase downsampling (LPD) to learn a shift-equivariant downsampling layer. Given an input feature map $\vx \in \sR^{C \times N}$, LPD spatially downsamples the input to produce an output in $\sR^{C \times \floor{N/2}}$ via

\begin{align}\label{eq:lpd}
    \text{LPD}(\vx)[c,n]= \vx[c,2n+k^{\star}]
    \triangleq  \poly{\vx}_{k^{\star}},
\end{align}

where $k^{\star}=\argmax_{k \in \{0,1\}}{p_\theta(\rvk=k|\vx)}$ and $\poly{\vx}_{k^{\star}}$ denotes the $k^{\star}$-th polyphase component. We model a conditional probability $p_\theta(\rvk|\vx)$ for selecting polyphase components, \ie, $\rvk$ denotes the  random variable of the polyphase indices. For 1D data, there are only two polyphase components.

Critically, {\it not all} $p_\theta$ lead to an equivariant downsampling layer. For example, $p_\theta(\rvk=0|\vx)=1$ results in the standard down-sampling which always returns values on even indices for 1D signals.  We will next examine which family of $p_\theta$ achieves a shift-equivariant downsampling layer.

{\noindent \bf Shift-permutation equivariance of $p_\theta$.}
Consider the example in~\figref{fig:pipeline}. We can see that a circular shift in the spatial domain induces a permutation in the polyphase components. Observe that the top-row of the polyphase component containing the blue circle and orange square are permuted to the second row when the input is circularly shifted. We now state this formally.

\begin{mdframed}[style=MyFrame,align=center]
\vspace{0.01cm}
\begin{lemma} Polyphase shift-permutation property
\label{lemma:poly_permute_prop.}
\bea\label{eq:poly_idx_permute}
\poly{\mT_N\vx}_k =  
\begin{cases}
      \poly{\vx}_1 & \text{if}~~k=0\\
      \mT_M\poly{\vx}_0 & \text{if}~~k=1
\end{cases}.
\eea
\end{lemma}
\end{mdframed}
\begin{proof}
\vspace{-0.3cm}
By definition, $\poly{\mT_N\vx}_k[n]$

\bea
&=& \mT_N\vx[(2n+k) {\;\tt mod\;} N]
=\vx[(2n+ k+1){\;\tt mod\;}N]\\
&=&
\hspace{-0.28cm}
\begin{cases}
       \vx[(2n+1){\;\tt mod\;}N] ~~~= \poly{\vx}_1 & \text{if}~~k=0\\
       \vx[(2(n+1)){\;\tt mod\;}N] = \mT_M\poly{\vx}_0 & \text{if}~~k=1
\end{cases}
\eea
\vspace{-0.2cm}
\end{proof}
\vspace{-0.2cm}
From Lemma~\ref{lemma:poly_permute_prop.}, we observe that to achieve an equivariant downsampling layer a spatially {\bf shifted input should lead to a permutation of the selection probability} (Claim~\ref{claim:lps_proof}). We note that $p_\theta$ is said to be \textit{shift-permutation-equivariant} if
\be\label{eq:shift_permute}
p_\theta(\rvk = \pi(k) |\mT_N \vx) =  p_\theta(\rvk=k|\vx),
\ee
where $\pi$ denotes a permutation on the polyphase indices, \ie, a ``swap'' of indices is characterized by $\pi(k)$, \ie, $\pi(0)=1$ and $\pi(1)=0$.

\begin{mdframed}[style=MyFrame,align=center]
\vspace{0.01cm}
\begin{claim}\label{claim:lps_proof}
If $p_\theta$ is shift-permutation-equivariant, defined in~\equref{eq:shift_permute}, then LPD defined in~\equref{eq:lpd} is a shift-equivariant downsampling layer.
\end{claim}
\end{mdframed}
\begin{proof}
\vspace{-0.3cm}
Let $\hat{\vx}\triangleq \mT_{N}\vx$ be a shifted version of $\vx \in \mathbb{R}^{N}$. Recall $\text{LPD}(\vx)$ and $\text{LPD}(\hat{\vx})$ are defined as:
\begin{align}
    \text{LPD}(\vx)\triangleq&\ \text{Poly}(\vx)_{k^{\star}},\ k^{\star}=\underset{k\in\{0,1\}}{\text{arg max}}\ p_{\theta}(\mathbf{k}=k|\vx),\\
    \text{LPD}(\hat{\vx})\triangleq&\ \text{Poly}(\hat{\vx})_{\hat{k}^{\star}},\ \hat{k}^{\star}=\underset{k\in\{0,1\}}{\text{arg max}}\ p_{\theta}(\mathbf{k}=k|\hat{\vx}).
\end{align}
From Lemma~\ref{lemma:poly_permute_prop.}, $\text{LPD}(\mT_{N}\vx)$ can be expressed as:
\begin{align}
\label{eq:c01_eq01}
    \text{LPD}(\mT_{N}\vx)=&\ \begin{cases}
                            \text{Poly}(\vx)_{1} &\ \text{if}~~\hat{k}^{\star}=0\\
                            \mT_{M}\text{Poly}(\vx)_{0} &\  \text{if}~~\hat{k}^{\star}=1
                            \end{cases}.
\end{align}
As $p_{\theta}$ is the shift-permutation-equivariant, 
\begin{align}
\label{eq:c01_eq02}
    \hat{k}^{\star}=&\ \pi(k^{\star})=1-k^{\star}.
\end{align}
Finally, combining~\equref{eq:c01_eq01} and~\equref{eq:c01_eq02},
\begin{align}
    \text{LPD}(\mT_{N}\vx)=&\
    \begin{cases}
        \text{Poly}(\vx)_{1}&\ \text{if}~~k^{\star}=1\\
        \mT_{M}\text{Poly}(\vx)_{0}&\ \text{if}~~k^{\star}=0
    \end{cases}
= \big((1-k^{\star})\mT_{M}+k^{\star}\mI\big) \cdot \text{LPD}(\vx),
\end{align}
showing that LPD satisfies the shift-equivariance definition reviewed in~\equref{eq:shift-eq-def}.
\end{proof}

Here, we parameterize $p_\theta$ with a small neural network. The exact construction of a shift-permutation equivariant deep-net architecture is deferred to~\secref{subsec:lps_design}. 
We next discuss how to train the distribution parameters $\theta$ in LPD. 

{\noindent \bf End-to-end training of LPD.} At training time, to incorporate stochasticity and compute gradients, we parameterize $p_\theta$ using
Gumbel Softmax~\cite{Jang_Gu_Poole_2017, Maddison_Mnih_Teh_2017}. To backpropagate gradients to $\theta$, we relax the selection of polyphase components as a convex combination, \ie,
\bea
\vy = \sum_{k} \vz_k \cdot \poly{\vx}_k, ~~\vz \sim p_{\theta}(\vk|\vx),
\eea
where $\vz$ corresponds to a selection variable, \ie, $\sum_k \vz_k = 1$ and $\vz_k \in [0,1]$. Note the slight abuse of notation as $p_{\theta}(\vk|\vx)$ denotes a probability over polyphase indices represented in a one-hot format. We further encourage the Gumbel Softmax to behave more like an argmax by decaying its temperature $\tau$ during training as recommended by~\citet{Jang_Gu_Poole_2017}. 

{\noindent \bf LPD generalizes APS.} A key advantage of LPS over APS is that it can learn from data, potentially leading to a better sampling scheme than a handcrafted one. Here, we show that APS is a special case of LPD. Therefore, LPD should perform at least as well as APS if parameters are trained well.

\begin{mdframed}[style=MyFrame,align=center]
\vspace{0.05cm}
\begin{claim}
APS is a special case of LPD, \ie, LPD can represent APS's selection rule.
\end{claim}
\end{mdframed}
\begin{proof}\vspace{-0.3cm}
Consider a parametrization of $p_\theta$ as follows,
\be p_\theta(\rvk=k|\vx)  = \frac{\exp{(\norm{\poly{\vx}_k})}}{\sum_{j} \exp(\norm{\poly{\vx}_j})}.
\label{eq:aps_prob}
\ee
As the exponential is a strictly increasing function we have 
\be
\argmax_k p_\theta(\rvk=k|\vx) = \argmax_k \norm{\poly{x}_k}.
\ee
\equref{eq:aps_prob} is a softmax with input $\norm{\poly{\vx}_k}$, as such a function exists, LPD generalizes APS.
\end{proof}

\subsection{Practical LPD Design}\label{subsec:lps_design}
We aim for a conditional distribution $p_\theta$ that is shift-permutation equivariant to obtain a shift-equivariant pooling layer. Let the conditional probability be modeled as:

\begin{align}\label{eq:p_theta_model}
    p_{\theta}(\rvk=k|\vx)\triangleq\ \frac{\exp[f_{\theta}(\poly{\vx}_{k})]}{\sum_{j}\exp[f_{\theta}(\poly{\vx}_{j})]},
\end{align}
where $f_{\theta}:\mathbb{R}^{C\times H'\times W'}\mapsto\mathbb{R}$ is a small network that extracts features from polyphase component $\poly{\vx}_k$. We first show that $p_{\theta}$ is shift-permutation equivariant if $f_{\theta}$ is shift invariant.

\begin{mdframed}[style=MyFrame,align=center]
\vspace{0.05cm}
\begin{claim}\label{claim:lpd_design}
In~\equref{eq:p_theta_model}, if $f_{\theta}$ is shift invariant then $p_\theta$ is shift-permutation equivariant (\equref{eq:shift_permute}).
\end{claim}
\end{mdframed}
\begin{proof}\vspace{-.2cm}
Denote a feature map $\vx$ and its shifted version $\hat{\vx}\triangleq \mT_{N}\vx$. By definition,
\begin{align}
    p_{\theta}(\rvk=\pi(k)|\mT_{N}\vx)=\ \frac{\exp(f_{\theta}(\poly{\mT_{N}\vx)_{\pi(k)}})}{\sum_{j}\exp(f_{\theta}(\poly{\mT_{N}\vx)_{j}})}.
\end{align}

\noindent With a shift-invariant $f_{\theta}$ and using Lemma~\ref{lemma:poly_permute_prop.},
\begin{align}
    f_{\theta}(\poly{\mT_{N}\vx}_{\pi(k)}) =& \ f_{\theta}(\mT_M\poly{\vx}_{k}) = f_{\theta}(\poly{\vx}_{k})\\
    \therefore\ p_{\theta}(\vk=\pi(k)|\mT_{N}\vx)=&\ \frac{\exp(\poly{\vx}_{k})}{\sum_{j}\exp(\rvk=\poly{\vx}_{j})} = \ p_{\theta}(\rvk=k|\vx)\qedhere
\end{align}
\end{proof}

\begin{figure}[t]
\centering
\includegraphics[width=0.9\columnwidth]{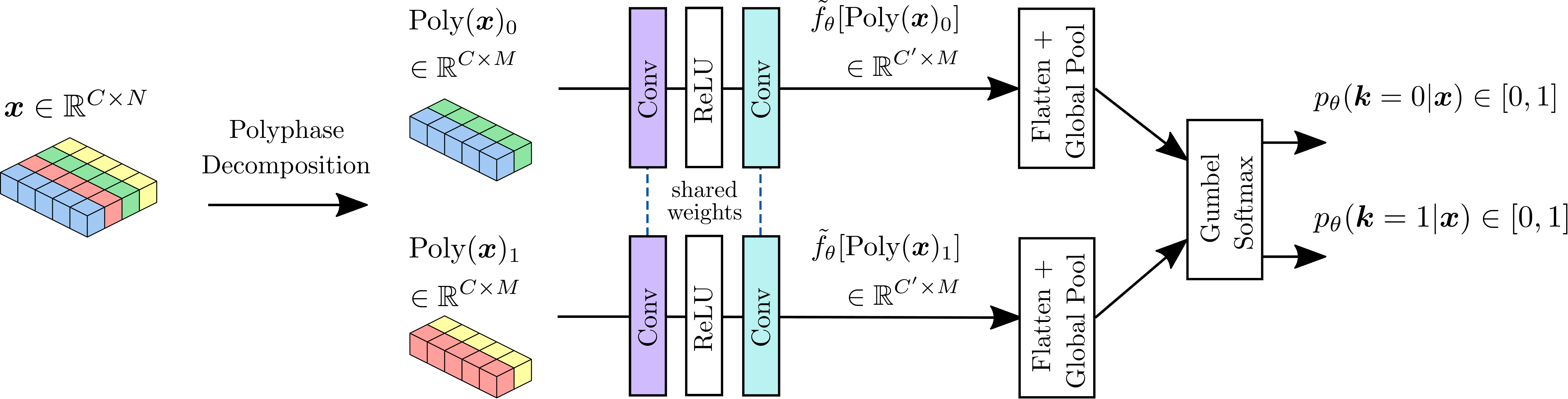}
\vspace{-0.1cm}
\caption{Proposed shift-permutation equivariant model.
By sharing weights across components and computing logits via global pooling, this neural network is shift-permutation equivariant.
To introduce stochasticity during training, we adopt the Gumbel Softmax for sampling.
}
\label{fig:conv_transformer_based}
\vspace{-0.3cm}
\end{figure}

Based on the result in~\clmref{claim:lpd_design}, we now present a convolution based meta-architecture that satisfies the shift-permutation property. The general design principle: share  parameters across polyphase indices, just as convolution achieves shift equivariance by sharing parameters, plus averaging over the spatial domain. An illustration of the proposed meta-architecture is shown in~\figref{fig:conv_transformer_based}.

{\noindent \bf Fully convolutional model.} Logits are extracted from the polyphase components via  fully-convolutional operations followed by averaging along the channel and the spatial domain. Following this, $f^{\text{conv}}_{\theta}$ is denoted as:
\begin{align}
    f^\text{conv}_{\theta}(\poly{\vx}_{k})\triangleq&\ \frac{1}{CM}\sum_{c,n}\tilde{f}^\text{conv}_{\theta}(\poly{\vx}_{k})[c,n], %
\end{align}
where $\tilde{f}^\text{conv}_{\theta}:\mathbb{R}^{C\times M}\mapsto \mathbb{R}^{C \times M}$ is a CNN model (without pooling layers) and $M=\floor{N/2}$. The shift equivariance property of $\tilde{f}^{\text{conv}}_{\theta}$ guarantees that $f^\text{conv}_{\theta}$ is shift-invariant due to the global pooling.

\subsection{Learnable Polyphase Upsampling (LPU)}\label{subsec:extension}
Beyond shift invariant models, we extend the  theory from downsampling to upsampling, which permits to design shift-equivariant models. The main idea is to place the features obtained after downsampling back to  their original spatial location. Given a feature map $\vy \in \sR^{C \times \floor{N/2}}$ downsampled via LPD from $\vx$, the upsampling layer outputs $\vu \in \sR^{C \times N}$ are defined as follows:
\bea\label{eq:lpu_def}
\poly{\vu}_{k^\star} = \begin{cases}
\vy,~~~k^\star = \argmax_{k \in \{0,1\}} p_\theta(\vk = k |\vx)\\
\mathbf{0},~~~\text{otherwise}.
\end{cases}
\eea
We name this layer learnable polyphase upsampling (LPU), \ie, $\text{LPU}(\vy , p_\theta) \triangleq \vu$. We now show that LPU and LPD achieve shift-equivariance.

\begin{mdframed}[style=MyFrame,align=center]
\vspace{0.05cm}

\begin{restatable}{claim}{vb}
If $p_{\theta}$ is shift-permutation equivariant, as defined in~\equref{eq:shift_permute}, then $\text{LPU}\circ \text{LPD}$ is shift-equivariant. 
\label{clm:vb}
\end{restatable}
\end{mdframed}
\begin{proof}\vspace{-0.3cm}
We prove this claim following definitions of LPU, LPD and Lemma~\ref{lemma:poly_permute_prop.}. The complete proof is deferred to Appendix~\secref{supp_sec:proof}. 
\end{proof}

{\bf End-to-end training of LPU.} As in downsampling, we also incorporate stochasticity via Gumbel-Softmax. To backpropagate gradients to $p_\theta$, we relax the hard selection into a convex combination,~\ie,
\bea
\poly{\vu}_{k} = \vz_k \cdot \vy, ~~~\vz\sim p_\theta(\vk|\vx).
\eea 

{\bf Anti-aliasing for upsampling.} While LPU provides a shift-equivariant upsampling scheme, it introduces zeros in the output which results in high-frequency components. This is known as aliasing in a multirate system~\cite{vetterli2014foundations}. To resolve this, following the classical solution, we apply a low-pass filter scaled by the upsampling factor after each LPU.

\section{Experiments}\label{sec:exp}

\begin{wraptable}[13]{r}{0.525\textwidth}
\vspace{-0.45cm}
\small
\caption{ResNet-18 (CIFAR10) top-1 accuracy and circular shift consistency. LPS outperforms all alternative pooling and anti-aliasing methods.
}
\vspace{-0.05cm}
\setlength{\tabcolsep}{4.5pt}
\resizebox{0.525\textwidth}{!}{
\begin{tabular}{ccccc}
\specialrule{.15em}{.05em}{.05em}
Method & Anti-Alias & Acc. $\uparrow$ & C-Cons. $\uparrow$ \\
\hline
Baseline & - & $91.44 \pm 0.33$ & $89.64 \pm 0.39$\\
APS & -& $94.07 \pm 0.28$ & $\bf 100 \pm 0.0$\\
LPS (Ours) & - & $\bf 94.45 \pm 0.05$ & $\bf 100 \pm 0.0$\\
\hline
LPF & Rect-$2$ & $93.1 \pm 0.17$ & $94.75 \pm 0.43$\\
APS & Rect-$2$ & $94.38 \pm 0.25$ & $\bf 100 \pm 0.0$\\
LPS (Ours)& Rect-$2$ & $\mathbf{94.69 \pm 0.06}$ & $\bf 100 \pm 0.0$\\
\hline
LPF & Tri-$3$ & $93.97 \pm 0.18$ & $96.74 \pm 0.2$ \\
APS & Tri-$3$ & $94.36 \pm 0.17$ & $\bf 100 \pm 0.0$\\
LPS (Ours)& Tri-$3$ & $\mathbf{94.8 \pm 0.14}$ & $\bf 100 \pm 0.0$\\
\hline
LPF & Bin-$5$ & $94.43 \pm 0.15$ & $98.34 \pm 0.15$ \\
APS & Bin-$5$ & $94.44 \pm 0.19$ & $\bf 100 \pm 0.0$\\
LPS (Ours)& Bin-$5$ & $\mathbf{94.49 \pm 0.1}$ & $\bf 100 \pm 0.0$\\
\hline
DDAC & Adapt-$3$ & $93.17 \pm 0.19$ & $95.13 \pm 0.15$\\
APS & Adapt-$3$ & $94.42 \pm 0.13$ & $\bf 100 \pm 0.0$\\
LPS (Ours)& Adapt-$3$ & $\mathbf{94.57 \pm 0.12}$ & $\bf 100 \pm 0.0$\\
\specialrule{.15em}{.05em}{.05em}
\end{tabular}
}
\label{tab:cifar10}
\end{wraptable}

We conduct experiments on image classification following prior works. We report on the same architectures and training setup. We report both the circular shift setup in APS~\cite{chaman2021truly} and the standard shift setup in LPF~\cite{zhang2019making}.

We also evaluate on semantic segmentation, considering the circular shift, inspired by APS, and the standard shift setup following DDAC~\cite{zou2020delving}. For circular shift settings, the theory exactly matches the experiment hence true equivariance is achieved. To our knowledge, this is the first truly shift equivariant model reported on PASCAL VOC.

\subsection{Image Classification (Circular Shift)}
\label{sec:image_class_circular}
{\noindent \bf Experiment \& implementation details.}

Following APS, all the evaluated pooling and anti-aliasing models use the ResNet-18~\cite{He_Zhang_Ren_Sun_2016} architecture with circular padding on CIFAR10~\cite{krizhevsky2009learning} and ImageNet~\cite{deng2009imagenet}. Anti-alias filters are applied after each downsampling layer following LPF~\cite{zhang2019making} and DDAC~\cite{zou2020delving}. We also replace downsampling layers with APS~\cite{chaman2021truly} and our proposed LPS layer. We provide more experimental details in Appendix~\secref{supp_sec:exp}.

\textbf{Evaluation metrics.} We report classification accuracy to quantify the model performance on the original dataset without any shifts. To evaluate shift-invariance, following APS, we report circular-consistency (C-Cons.) which computes the average percentage of predicted labels that are equal under two different circular shifts, \ie,
\be\label{eq:c_cons}
    \hat{\vy}(\text{Circ. Shift}_{h1,w1}(I)) = \hat{\vy}(\text{Circ. Shift}_{h2,w2}(I)),
\ee
where $\hat{\vy}(I)$ denotes the predicted label for an input image $I$ and $h1,w1,h2,w2$ are uniformly sampled from $0$ to $32$. We report the average over five random seeds.

{\noindent \bf CIFAR10 results.}
\tabref{tab:cifar10} shows the classification accuracy and circular consistency on CIFAR10. We report the mean and standard deviation over five runs with different random initialization of the ResNet-18 model. We observe that the proposed LPS improves classification accuracy over all baselines while achieving 100\% circular consistency. In addition to attaining perfect shift consistency, we observe that combining anti-aliasing with LPS further improves performance.

\textbf{ImageNet results.}
We conduct experiments on ImageNet with circular shift using ResNet-18. In~\tabref{tab:imagenet_circular}, we compare with APS's best model using a box filter (Rectangle-2), as reported by~\citet{chaman2021truly}. While both APS and LPS achieve 100\% circular consistency, our proposed LPS improves on classification accuracy in all scenarios, highlighting its advantages. 

\begin{table}[!ht]
\vspace{-0.25 cm}
\centering
\begin{minipage}[t]{.45\textwidth}
\setlength{\tabcolsep}{5pt}
\small
\captionof{table}{
ResNet-18 top-1 classification accuracy and circular consistency on ImageNet.
}
\begin{tabular}{cccc}
\specialrule{.15em}{.05em}{.05em}
Method & Anti-Alias & Acc.\ $\uparrow$ & C-Cons.\ $\uparrow$ \\
\hline
Baseline & - & $64.88$ & $80.39$\\
APS & - & $67.05$ & $\bf 100$\\
LPS (Ours) & - & $\mathbf{67.39}$ & $\bf 100$\\
\hline
LPF & Rect-$2$ & $67.03$ & $84.35$\\
APS & Rect-$2$ & $67.6$ & $\bf 100$\\
LPS (Ours) & Rect-$2$ & $\mathbf{68.45}$ & $\bf 100$\\
\hline
DDAC & Adapt-$3$ & $67.6$ & $77.23$\\
APS & Adapt-$3$ & $69.02$ & $\mathbf{100}$\\
LPS (Ours) & Adapt-$3$ & $\mathbf{69.11}$ & $\mathbf{100}$\\
\specialrule{.15em}{.05em}{.05em}
\end{tabular}
\label{tab:imagenet_circular}
\end{minipage}
\hspace{0.4cm}
\begin{minipage}[t]{.5\textwidth}
\renewcommand{\arraystretch}{0.99}
\setlength{\tabcolsep}{2pt}
\centering
\small
\captionof{table}{
ResNet-50/101 top-1 classification accuracy and shift consistency on ImageNet.
}
\begin{tabular}{ccccc}
\specialrule{.15em}{.05em}{.05em}
Method & Anti-Alias &Model & Acc.\ $\uparrow$ & S-Cons.\ $\uparrow$ \\
\hline
Baseline & - & ResNet-50 & $76.16$ & $89.2$\\
LPF & Tri-$3$ &ResNet-50 & $76.83$ & $90.91$\\
LPS (Ours) & Tri-$3$ &ResNet-50 & $\mathbf{77.14}$ & $\mathbf{91.43}$\\
\hline
Baseline & - & ResNet-101 & $77.7$ & $90.6$\\
LPF & Tri-$3$ & ResNet-101 & $78.4$ & $91.6$\\
LPS & Tri-$3$ & ResNet-101 & $\mathbf{78.51}$ & $\mathbf{91.69}$\\
\hline
DDAC & Adapt-$3$ & ResNet-101 & $\bf 79.0$ & $91.8$\\
DDAC$^{*}$ & Adapt-$3$ & ResNet-101 & $78.64$ & $91.83$\\
LPS (Ours) & Adapt-$3$ & ResNet-101 & $78.8$ & $\mathbf{92.4}$\\
\specialrule{.15em}{.05em}{.05em}
\end{tabular}
\label{tab:imagenet_standard}
\end{minipage}
\vspace{-0.35cm}
\end{table}

\subsection{Image Classification (Standard Shift)}\label{subsec:image_class_shift}
{\noindent \bf Experiment \& implementation details.}
To directly compare with results from LPF and DDAC, we conduct experiments on ImageNet using the ResNet-50 and ResNet-101 architectures following their setting, \ie, training with standard shifts augmentation and using convolution layers with zero-padding. 

{\noindent \bf Evaluation metrics.}
Shift consistency (S-Cons.) computes the average percentage of
\be
 \hat{\vy}(\text{Shift}_{h1,w1}(I)) = \hat{\vy}(\text{Shift}_{h2,w2}(I)),
\ee

where $h1,w1,h2,w2$ are uniformly sampled from the interval $\{0, \dots, 32\}$. To avoid padding at the boundary, following LPF~\cite{zhang2019making}, we perform a shift on an image then crop its center $224 \times 224$ region. We note that, due to the change in content at the boundary, perfect shift consistency is not guaranteed. 

{\noindent \bf ImageNet results.}
In~\tabref{tab:imagenet_standard}, we compare to the best anti-aliasing result as reported in LPF, DDAC and DDAC$^{*}$ which is trained from the authors' released code using hyperparameters specified in the repository. Note, in standard shift settings LPS no longer achieves true shift-invariance due to padding at the boundaries. Despite this gap from the theory, LPS achieves improvements in both performance and shift-consistency over the baselines. When compared to LPF, both ResNet-50 and ResNet-101 architecture achieved improved classification accuracy and shift-consistency. When compared to DDAC, LPS achieves comparable accuracy with higher shift-consistency. 

\subsection{Trainable Parameters and Inference Time}
While LPD is a data-driven downsampling layer, we show that the additional trainable parameters introduced by it are marginal with respect to the classification architecture. \tabref{tab:supp_train_pars} shows the number of trainable parameters required by the ResNet-101 models.

For each method, we report the \textit{absolute} number of trainable parameters, which includes both classifier and learnable pooling weights. We also include the \textit{relative} number of trainable parameters, which only considers the learnable pooling weights and the percentage it represents with respect to the default ResNet-101 architecture weights. 

For comparison purposes, we also include the inference time required by each model to evaluate their computational overhead. Mean and standard deviation of the inference time is computed for each method on $100$ batches of size $32$. Following ImageNet default settings, the image dimensions corrrespond to $224 \times 224 \times 3$.

Results show our proposed LPD method introduces approximately $1\%$ additional trainable parameters on the ResNet-101 architecture, and increases the inference time roughly by $14.89$ ms over the LPF anti-aliasing method (the less computationally expensive of the evaluated techniques). On the other hand, most of the overhead comes from DDAC, which increases the number of trainable parameters by approximately $4\%$ and the inference time by approximately $55.97$ ms. Overall, our comparison shows that, by equipping a classifier with LPD layers, the computational overhead is almost trivial.

Despite increasing the number of trainable parameters, we empirically show that our LPD approach outperforms classifiers with significantly more parameters. Please refer to \secref{sec:supp_comparison} for additional experiments comparing the performance of our ResNet-101 + LPD model against the much larger ResNet-152 classifier.

\begin{table*}[t]
\begin{center}
\caption{\label{tab:supp_train_pars} Number of trainable parameters and inference time required by ResNet-101 using our proposed LPD layer and alternative methods. \textit{Absolute} corresponds to the total number of trainable parameters, including classifier and pooling layers. \textit{Relative} corresponds to the parameters of the pooling layers only. Inference time statistics computed on $100$ batches with $32$ images of size $224 \times 224 \times 3$ each.}
\resizebox{0.9\textwidth}{!}{
\begin{tabular}{c|c|c|c|c|c}
\specialrule{.15em}{.05em}{.05em} 
Pooling & Anti-alias & \multicolumn{2}{c|}{Trainable Parameters} & \multicolumn{2}{c}{Inference Time}\\
\midrule
\multicolumn{2}{c|}{} & Absolute & Relative & Mean (ms) & Std (ms)\\
\midrule
LPD & LPF (Tri-3) & $42,966,730$ & $446,080\ (1.05 \%)$ & $83.61
$ & $0.19$\\
LPD & DDAC (Adapt-3) & $44,751,034$ & $2,230,384\ (5.24 \%)$ & $130.69
$ & $0.2$\\
ResNet-101 (Default) & DDAC (Adapt-3) & $44,304,954$ & $1,784,304\ (4.2 \%)$ & $124.69
$ & $0.25$\\
APS & LPF (Tri-3) & $42,520,650$ & $0$ & $77.68
$ & $0.22$\\
ResNet-101 (Default) & LPF (Tri-3) & $42,520,650$ & $0$ & $68.72
$ & $0.4$\\
\midrule
\end{tabular}}
\end{center}
\vspace{-0.7\baselineskip}
\end{table*}

\begin{figure}[t]
\centering
\includegraphics[width=\columnwidth,
trim={0 1.2cm 0 0},clip
]{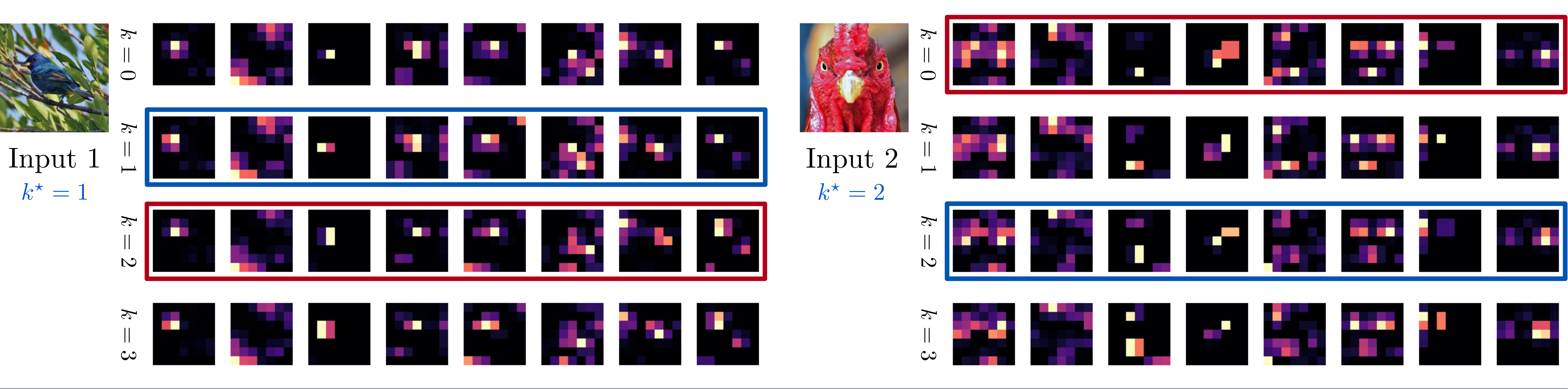}
\vspace{-0.2cm}
\small
\caption{ResNet-50 LPS activation maps (ImageNet). We show all four polyphase components of the fourth layer (first $8$ channels). The  component selected by {\color{mybb} LPS is boxed in blue}, while that with the {\color{myrr} largest $\ell_{2}$-norm is boxed in red.} This indicates LPS did not learn to reproduce APS.} 
\label{fig:activations}
\vspace{-0.4cm}
\end{figure}

{\noindent \bf LPD learns sampling schemes different from APS.}
To further analyze LPD, we replace all the LPD layers with APS for a ResNet-101 model trained on ImageNet. We observe a critical drop in top-1 classification accuracy from $78.8\%$ to $0.1\%$, indicating that LPD did not learn a downsampling scheme equivalent to APS. We also counted how many times (across all layers) LPD selects the max-norm. On the ImageNet validation set, LPD  selects the max $\ell_{2}$-norm polyphase component only $20.57\%$ of the time. These show LPD learned a selection rule that differs from the handcrafted APS.

{\noindent \bf Qualitative study on LPD.}
In~\figref{fig:activations} we show the selected activations, at the fourth layer, of a ResNet-50 model with LPD. Each column describes the first $8$ channels of the four possible polyphase components $k\in \{0,\dots,3\}$. The component selected by LPD, denoted as $k^{\star}$, is boxed in blue. For comparison purposes, we also boxed the component that maximizes the $\ell_{2}$-norm in red. We observe that LPD is distinct from APS as they select a different set of polyphase components. However, we did not observe a specific pattern that can explain LPD's selection rule.

\subsection{Semantic Segmentation (Circular Shift)}\label{subsec:segm_circular}
{\noindent \bf Experiment \& implementation details.}
We evaluate LPS's down/upsampling layers on semantic segmentation. As in DDAC~\cite{zou2020delving}, we evaluate using the PASCAL VOC~\cite{everingham2015pascal} dataset. Following DDAC, we use DeepLabV3+~\cite{chen2018encoder} as our baseline model. We use the ResNet-18 backbone pre-trained on the ImageNet (circular shift) reported in~\secref{sec:image_class_circular}. We experiment with using only the LPD backbone and the full LPS, \ie, both LPD and LPU.

We also evaluate the performance using APS, which corresponds to a hand-crafted downsampling scheme, in combination with the default bilinear interpolation strategy from DeepLabV3+. Note that, while our LPS approach consists of both shift equivariant down and upsampling schemes (LPD and LPU, respectively), APS only operates on the downsampling process. Thus, the latter does not guarantee a circularly shift equivariant segmentation.

{\noindent \bf Evaluation metric.} We report mean intersection over union (mIoU) to evaluate segmentation performance. To quantify circular-equivariance, we report mean Average Segmentation Circular Consistency (mASCC) which computes the average percentage of predicted (per-pixel) labels that remained the same under two different circular shifts. \Ie, a shifted image is passed to a model to make a segmentation prediction. This prediction is then ``unshifted'' for comparison. We report five random shift pairs for each image.

{\noindent \bf Results.}
We report the results for PASCAL VOC in~\tabref{tab:semantic_circular}. Overall, we observe that LPD only and LPS achieve comparable results to DDAC and APS in mIoU. Notably, LPS achieves 100\% mASCC, matching the theory. This confirms that both the proposed LPD and LPU layers are necessary and are able to learn effective down/up sampling schemes for semantic segmentation.

\subsection{Semantic Segmentation (Standard Shift)}
{\noindent \bf Experiment \& implementation details.}
For the standard shift setting, we directly follow the experimental setup from DDAC. We use DeepLabV3+ with a ResNet-101 backbone pre-trained on ImageNet as reported in~\secref{subsec:image_class_shift}.

{\noindent \bf Evaluation metric.} To quantify the shift-equivariance, following DDAC, we report the mean Average Semantic Segmentation Consistency (mASSC) which is a linear-shift version of mASCC described in~\secref{subsec:segm_circular} except boundary pixels are ignored.

\begin{table}[t]
\centering
\begin{minipage}{0.48\textwidth}
\small
\caption{\label{tab:semantic_circular}Semantic segmentation with circular shifts (ResNet-18 with DeepLabV3+) on PASCAL VOC.}
\begin{center}
\resizebox{1\columnwidth}{!}{
\begin{tabular}{cc  cc}
\specialrule{.15em}{.05em}{.05em}
Method & Anti-Alias & mIoU$ \uparrow$ & mASCC$ \uparrow$\\
\hline
DeepLabV3+ & - & 70.03 & 95.42\\
LPF & Rect-2 & 71.02 & 96.03\\
DDAC & Adapt-$3$ & 72.28 & 95.98\\
APS & Adapt-$3$ & 72.37 & 96.70\\
LPD only & Adapt-$3$ & \bf 72.47 & 96.23\\
LPS (Ours) & Adapt-$3$ & 72.37 & \bf 100 \\
\specialrule{.15em}{.05em}{.05em}
\end{tabular}
}
\end{center}
\end{minipage}\hfill
\begin{minipage}{0.48\textwidth}
\small
\caption{\label{tab:semantic}Semantic segmentation with standard shifts (ResNet-101 with DeepLabV3+) on PASCAL VOC. Results above the line are from DDAC's paper.}
\begin{center}
\resizebox{1\columnwidth}{!}{
\centering
\begin{tabular}{cc  cc}
\specialrule{.15em}{.05em}{.05em}
Method & Anti-Alias & mIoU $\uparrow$ & mASSC $\uparrow$\\
\hline
DeepLabV3+ & - & 78.5 & 95.5 \\
LPF & Tri-$3$ & 79.4 &  95.9 \\
DDAC & Adapt-$3$ & 80.3 &  96.3\\
\hline
DDAC* & Adapt-$3$ & 80.31 &  97.83\\
LPS (Ours) & Adapt-$3$ & \bf 80.43 & \bf 97.98\\
\specialrule{.15em}{.05em}{.05em}
\end{tabular}
}
\end{center}
\end{minipage}
\vspace{-0.35cm}
\end{table}

{\noindent \bf Results.}
In~\tabref{tab:semantic}, we compare mIoU and mASSC of LPS to various baselines. We observe that LPS achieves improvements in mIoU and consistency when compared to $\text{DDAC}^*$. We note that DDAC~\cite{zou2020delving} did not release their code for mASSC. For a fair comparison, we report the performance of their released checkpoint using our implementation of mASSC, indicated with $\text{DDAC}{^*}$.
Despite the gap in theory and practice due to non-circular padding at the boundary, our experiments show LPS remains an effective approach to improve both shift consistency and model performance.

\section{Conclusion}\label{sec:conc}
We propose learnable polyphase sampling (LPS), a pair of shift-equivariant down/upsampling layers. LPS's design theoretically guarantees circular shift-invariance and equivariance while being end-to-end trainable. Additionally, LPS retains superior consistency on standard shifts where theoretical assumptions are broken at image boundaries. Finally, LPS captures a richer family of shift-invariant/equivariant functions than APS. Through extensive experiments on image classification and semantic segmentation, we demonstrate that LPS is on-par with/exceeds APS, LPF and DDAC in terms of model performance and consistency. 

{\small
{\bf Acknowledgments:} 
We thank Greg Shakhnarovich \&  \href{https://pals.ttic.edu/}{PALS} at TTI-Chicago for the thoughtful discussions and computation resources. This work is supported in part by NSF under Grants 1718221, 2008387, 2045586, 2106825, MRI 1725729, NIFA award 2020-67021-32799, and funding by PPG Industries, Inc. We thank NVIDIA for providing a GPU. 
}

\bibliographystyle{abbrvnat}
\bibliography{ref_equivariant.bib}

\newpage
\appendix
\onecolumn

\newcommand{\beginsupplementary}{
    \setcounter{section}{0}
	\renewcommand{\thesection}{A\arabic{section}}
	\renewcommand{\thesubsection}{\thesection.\arabic{subsection}}

	\renewcommand{\thetable}{A\arabic{table}}%
	\setcounter{table}{0}

	\renewcommand{\thefigure}{A\arabic{figure}}%
	\setcounter{figure}{0}
}
\newcommand{\suptitle}{Appendix for:\\\papertitle}

\newcommand{\toptitlebar}{
	\hrule height 4pt
	\vskip 0.25in
	\vskip -\parskip%
}
\newcommand{\bottomtitlebar}{
	\vskip 0.29in
	\vskip -\parskip
	\hrule height 1pt
	\vskip 0.09in%
}

\newcommand{\maketitlesupp}{
	\vbox{
		\hsize\textwidth
		\linewidth\hsize
		\vskip 0.1in
		\toptitlebar
		\centering
		{\LARGE\bf \suptitle \par}
		\bottomtitlebar
		\vskip 0.3in
	}
}

\beginsupplementary
{\bf \LARGE Appendix}
\vspace{0.1cm}

The appendix is organized as follows:
\begin{itemize}
\item In~\secref{supp_sec:proof}, we provide the complete proof of~\clmref{clm:vb}.
\item In~\secref{supp_sec:result}, we provide an additional ablation study and qualitative experimental results.
\item In~\secref{supp_sec:impl}, we provide an overview of the attached code and showcase the ease of incorporating LPS into a deep-net to achieve a shift-invariance/equivariance network. Unit tests and end-to-end tests empirically validating the theory are also provided.
\item In~\secref{supp_sec:exp}, we provide additional experimental details,~\eg, model architecture, number of trainable parameters, hyperparameters and baseline implementations.
\end{itemize}

\section{Proof of~\clmref{clm:vb}} \label{supp_sec:proof}
\begin{mdframed}[style=MyFrame]
\vb*
\end{mdframed}

\begin{proof}\vspace{-0.3cm}
Let $\vx$ be a feature map, $\hat{\vx}\triangleq \mT_{N}\vx$ its shifted version, and $k^\star = \argmax_{k \in \{0,1\}} p_\theta(\vk = k |\vx)$. By definition, $\vu \triangleq LPU \circ LPD (\vx)$ can be seen as masking out the components beside $k^\star$ from $\vx$:
\bea
\poly{\vu}_{j} = \begin{cases}
\poly{\vx}_{j}, &~~~j = k^\star\\
\mathbf{0}, &~~~\text{otherwise}.
\end{cases}
\eea
Let $\hat\vu \triangleq \text{LPU} \circ \text{LPD} (\hat\vx)$ and $\hat{k}^\star = \argmax_{k \in \{0,1\}} p_\theta(\vk = k |\hat\vx)$ then 
\bea
\poly{\hat\vu}_{j} = \begin{cases}
\poly{\hat\vx}_{j}, &~~~j = \hat{k}^\star\\
\mathbf{0}, &~~~\text{otherwise}.
\end{cases}
\eea
Using Lemma~\ref{lemma:poly_permute_prop.} on $\hat{\vx}$, when $j = 0$
\bea\label{eq:after_lemma1_j=0}
\poly{\hat\vu}_{j} = \begin{cases}
\poly{\vx}_{\pi(j)}, &~~~ \pi(j) = \hat{k}^\star\\
\mathbf{0},
&~~~\text{otherwise}.
\end{cases}
\eea

As $p_\theta$ is shift-permutation equivariant, therefore $\hat{k}^\star = \pi(k^\star)$. Substituting into~\equref{eq:after_lemma1_j=0}, 
\bea\label{eq:after_lemma1_j=0_2}
\poly{\hat\vu}_{j} = \begin{cases}
\poly{\vx}_{\pi(j)}, &~~~\pi(j) = \pi(k^\star)\\
\mathbf{0},
&~~~\text{otherwise}.
\end{cases}
\eea

Similarly, when $j = 1$ and let $M=\floor{N/2}$
\bea\label{eq:after_lemma1_j=1}
\mT_M\poly{\hat\vu}_{j} = \begin{cases}
\mT_M\poly{\vx}_{\pi(j)}, &~~~\pi(j) = \pi(k^\star)\\
\mathbf{0},
&~~~\text{otherwise}.
\end{cases}
\eea

Finally, combining~\equref{eq:after_lemma1_j=0_2} and~\equref{eq:after_lemma1_j=1} then using
Lemma~\ref{lemma:poly_permute_prop.} on $\hat\vu$,
\bea
\mT_N\vu = \text{LPU} \circ \text{LPD}(\mT_N\vx),
\eea
which proves the claim.
\end{proof}

\section{Additional Experimental Results}\label{supp_sec:result}

{\noindent\bf Ablation Study on Gumbel Softmax.} 
We further analyze the effect of sampling with Gumbel-Softmax during training. We compare LPS (ResNet-18) results using a $3$-tap antialiasing filter (Tri-$3$) and trained using a Gumbel-Softmax, against its alternative version trained using a standard Softmax without sampling. The remaining training attributes, including the $\tau$ annealing schedule, remain unaltered. \tabref{tab:supp_cifar_gumbel} compares both scenarios. While perfect circular shift consistency is obtained by design in both cases, the top-1 classification accuracy of the model trained without sampling ($93.22\%$) is significantly lower than our proposed training approach ($94.8\%$). By incorporating stochasticity, the model performance improves.

\begin{table}[ht]
\centering
\caption{Ablation study on the effect of Gumbel-softmax sampling.
}
\begin{tabular}{ccccc}
\specialrule{.15em}{.05em}{.05em}
Method & Anti-Alias & Sampling & Acc. $\uparrow$ & C-Cons. $\uparrow$ \\
\hline
\makecell{LPS (Softmax)}& Tri-$3$ & \xmark & $93.22 \pm 0.13$ & $\bf 100 \pm 0.0$\\
\makecell{LPS (Gumbel-softmax)}& Tri-$3$ & \cmark & $\mathbf{94.8 \pm 0.14}$ & $\bf 100 \pm 0.0$\\
\specialrule{.15em}{.05em}{.05em}
\end{tabular}
\vspace{-0.2cm}
\vspace{-0.2cm}
\label{tab:supp_cifar_gumbel}
\end{table}

\begin{figure}[H]
\centering
\includegraphics[width=1\columnwidth]{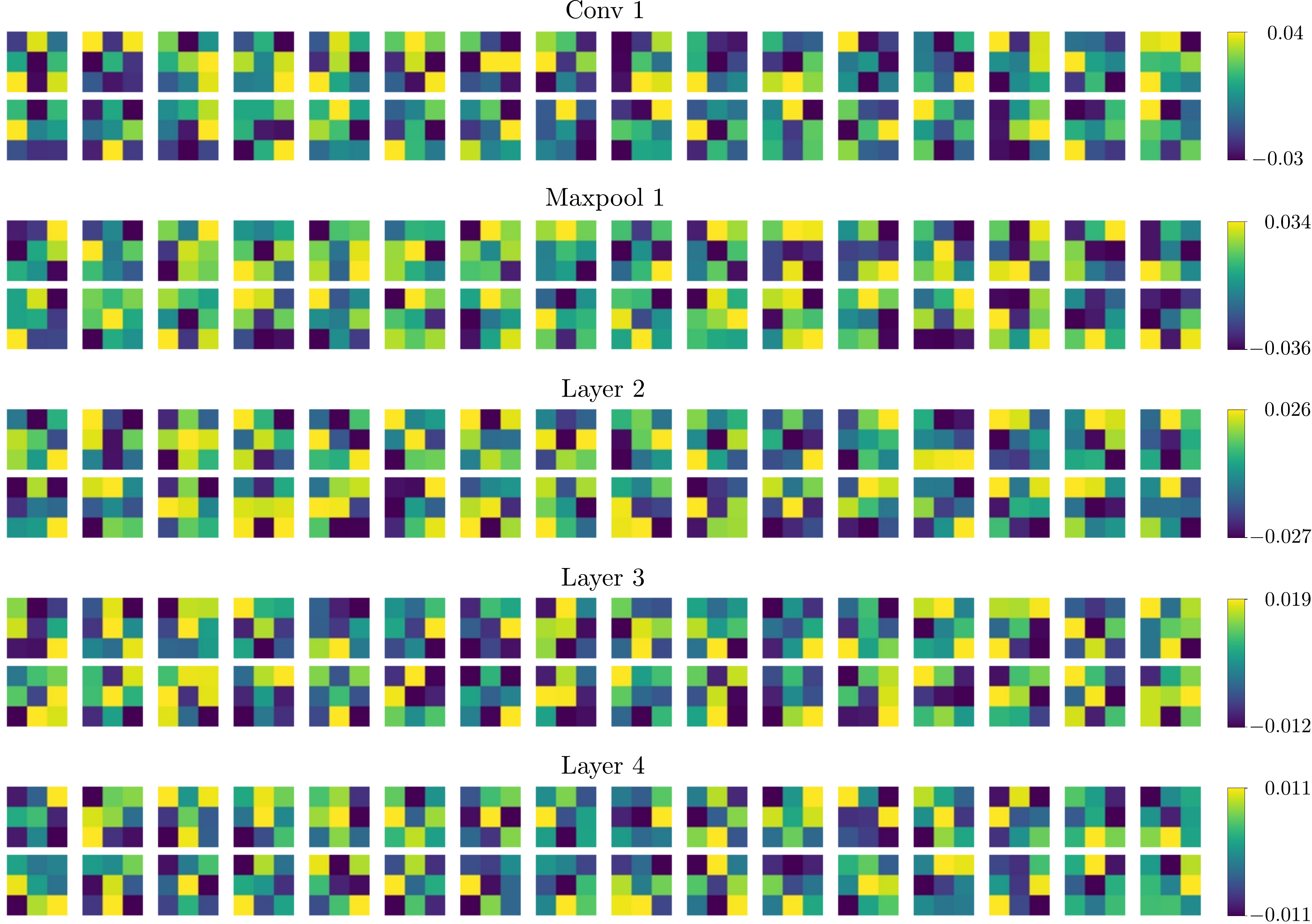}
\vspace{-0.1 cm}
\caption{Convolutional weights learned by LPD layers at different ResNet-50 (ImageNet) pooling levels.}
\label{fig:weights}
\end{figure}

{\noindent \bf LPS Filter Visualization.}
We provide visualizations of the convolution weights.~\figref{fig:weights} shows a subset of the convolutional kernels used to select polyphase components at each layer of a  ResNet-50 (LPS).

{\noindent \bf Qualitative Results for Semantic Segmentation.}
In~\figref{fig:semanctic_quan} to~\figref{fig:semanctic_quan_12}, we provide a sampling of output masks on PASCAL VOC that were predicted by DDAC and our proposed LPS on linearly shifted inputs. Identical random shifts were applied for both DDAC and LPS. We observe that LPS predicts smoother object contours and maintains a better consistency across shifts when compared to DDAC. Regions where our shift consistency property showed significantly different results in comparison to baseline are highlighted by a red circle.

\begin{minipage}{1\textwidth}
\centering
\textbf{Input}\\
\includegraphics[width=0.85\linewidth,trim={0 0 73cm 0},clip]{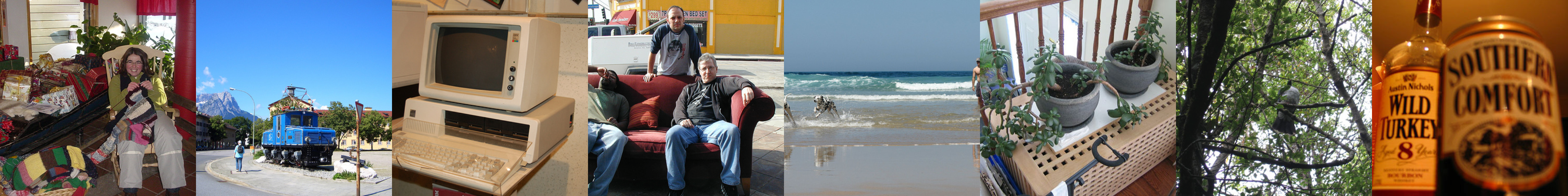}\\
\textbf{DDAC on Input}\\
\includegraphics[width=0.85\linewidth,trim={0 0 73cm 0},clip]{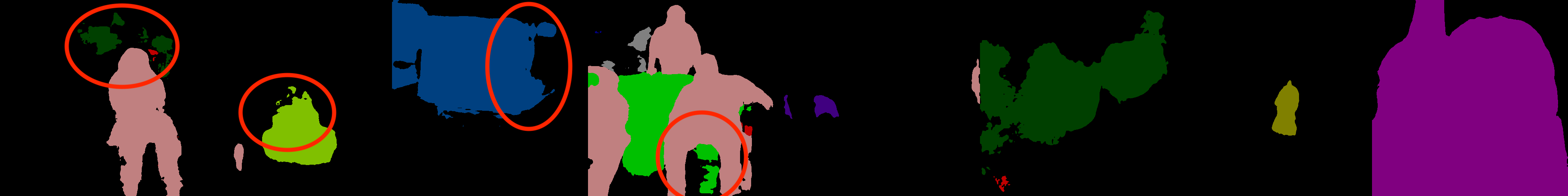}\\
\textbf{DDAC on Shifted Input}\\
\includegraphics[width=0.85\linewidth,trim={0 0 73cm 0},clip]{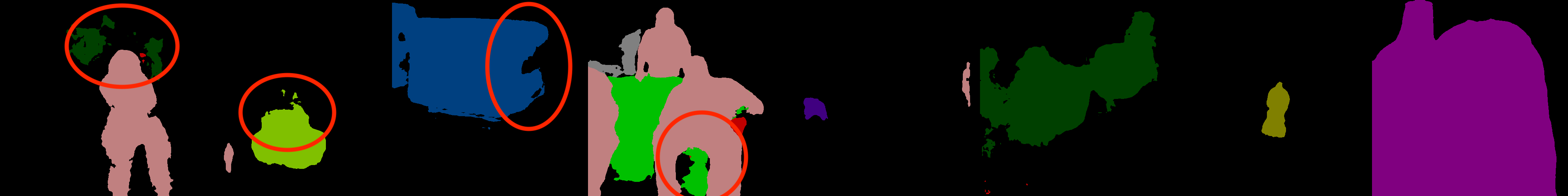}\\
\textbf{LPS on Input}\\
\includegraphics[width=0.85\linewidth,trim={0 0 73cm 0},clip]{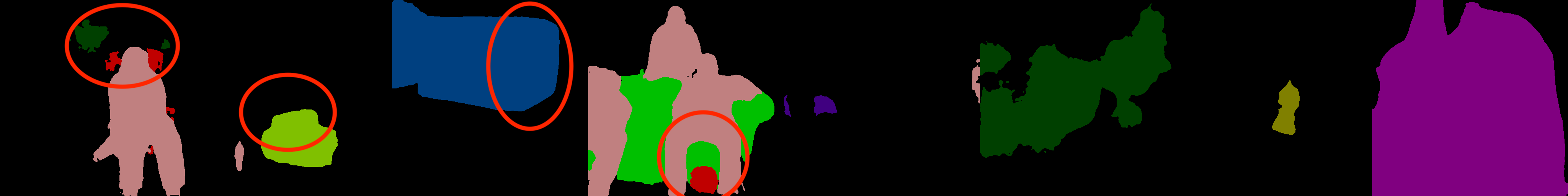}\\
\textbf{LPS on Shifted Input}\\
\includegraphics[width=0.85\linewidth,trim={0 0 73cm 0},clip]{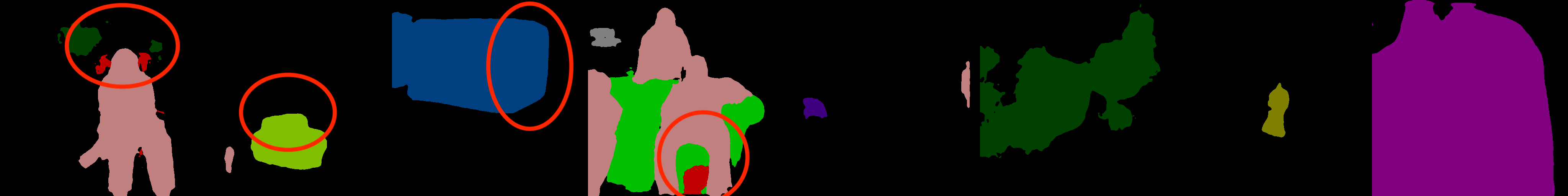}
\vspace{-0.2cm}
\captionof{figure}{Qualitative comparison on PASCAL VOC ResNet-101 with DeepLabV3 architecture. Regions where our proposed network showed significant improvements under linear shifts are highlighted with a red circle.}
\label{fig:semanctic_quan}
\vspace{-0.2cm}
\end{minipage}

\begin{minipage}{1\textwidth}
\centering
\textbf{Input}\\
\includegraphics[width=0.2125\linewidth]{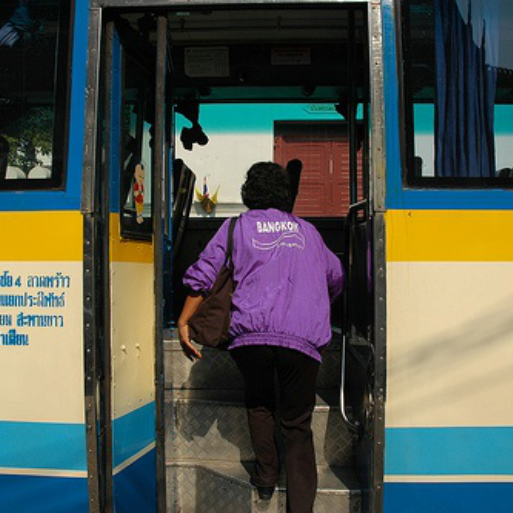}\includegraphics[width=0.2125\linewidth]{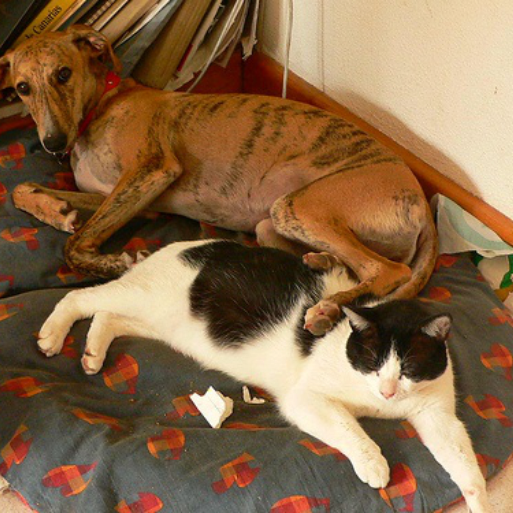}\includegraphics[width=0.2125\linewidth]{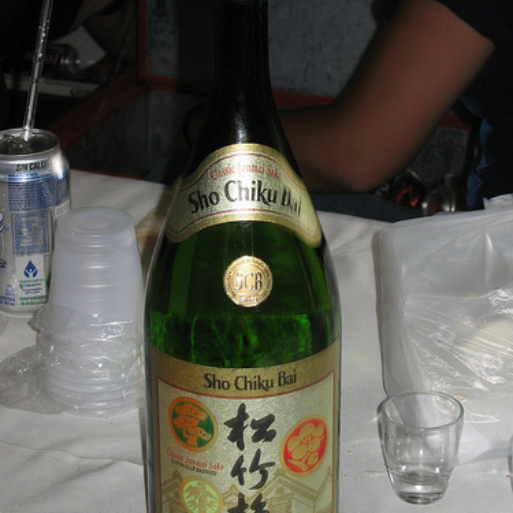}\includegraphics[width=0.2125\linewidth]{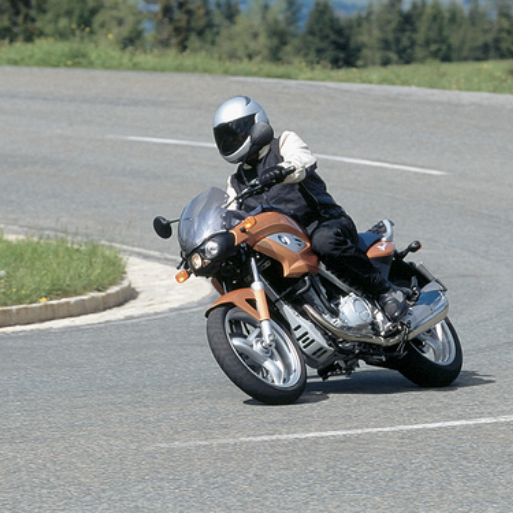}\\
\textbf{DDAC on Input}\\
\includegraphics[width=0.2125\linewidth]{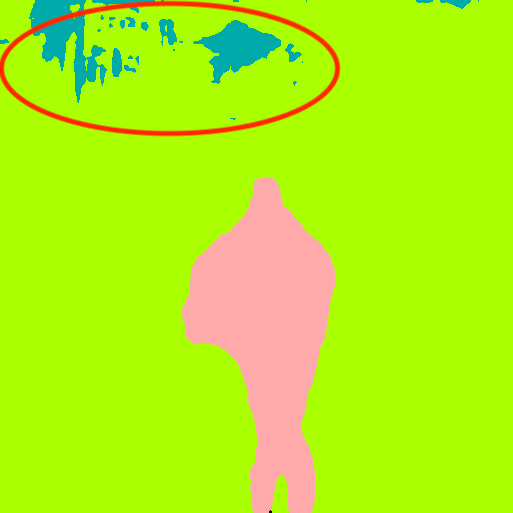}\includegraphics[width=0.2125\linewidth]{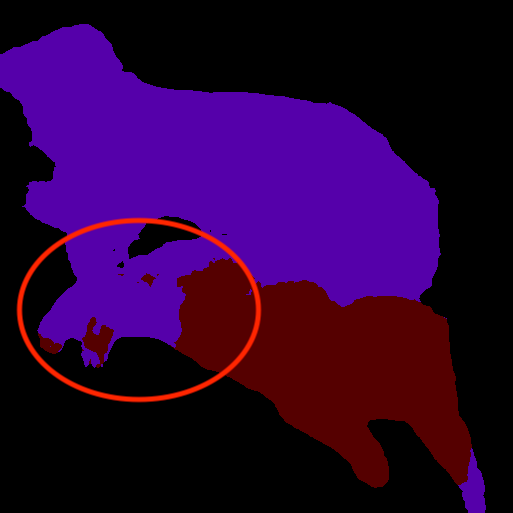}\includegraphics[width=0.2125\linewidth]{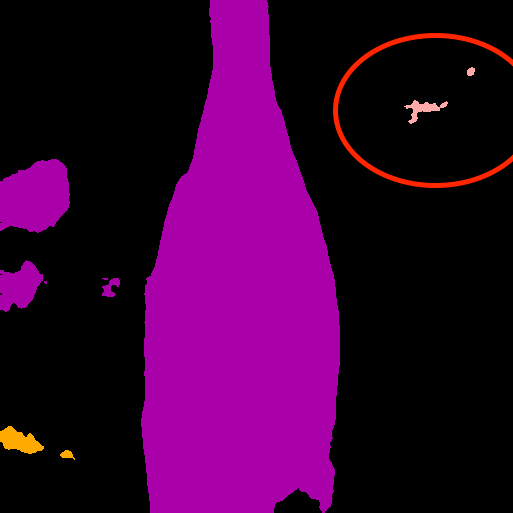}\includegraphics[width=0.2125\linewidth]{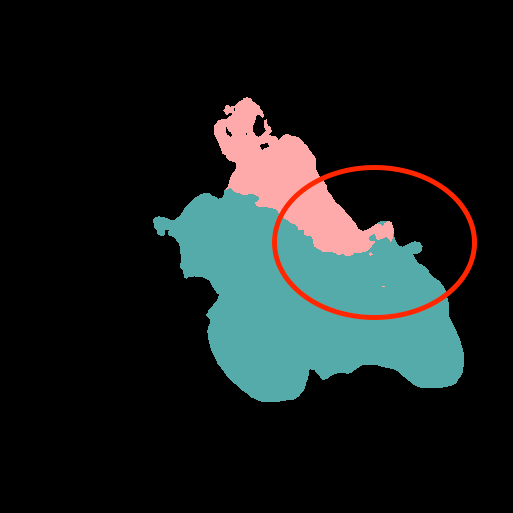}\\
\textbf{DDAC on Shifted Input}\\
\includegraphics[width=0.2125\linewidth]{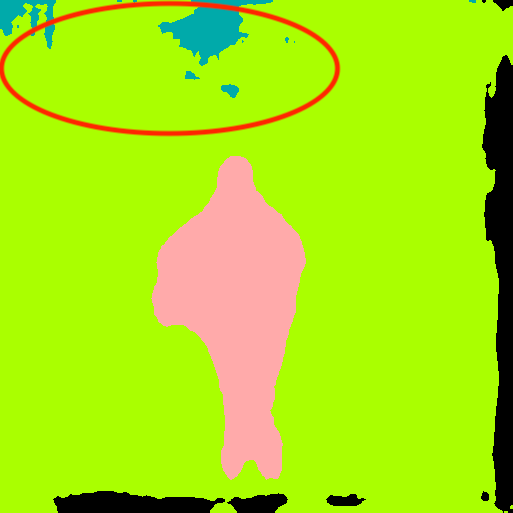}\includegraphics[width=0.2125\linewidth]{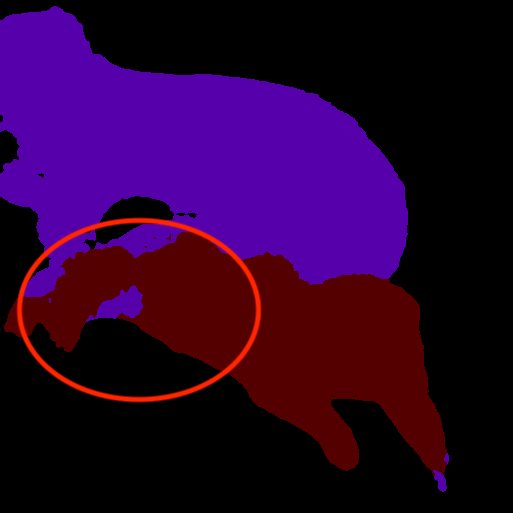}\includegraphics[width=0.2125\linewidth]{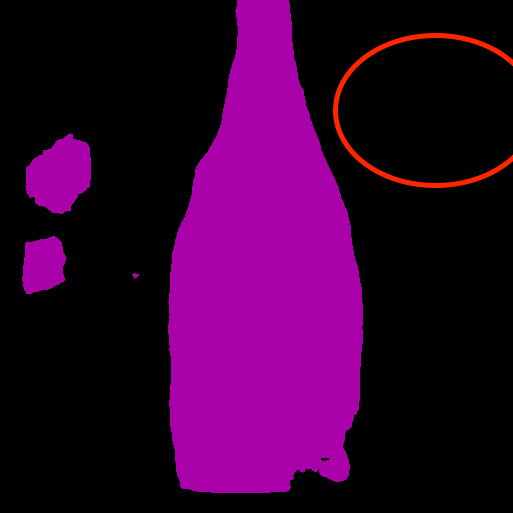}\includegraphics[width=0.2125\linewidth]{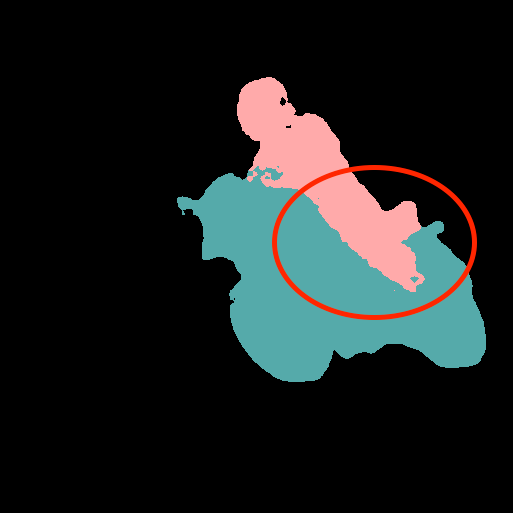}\\
\textbf{LPS on Input}\\
\includegraphics[width=0.2125\linewidth]{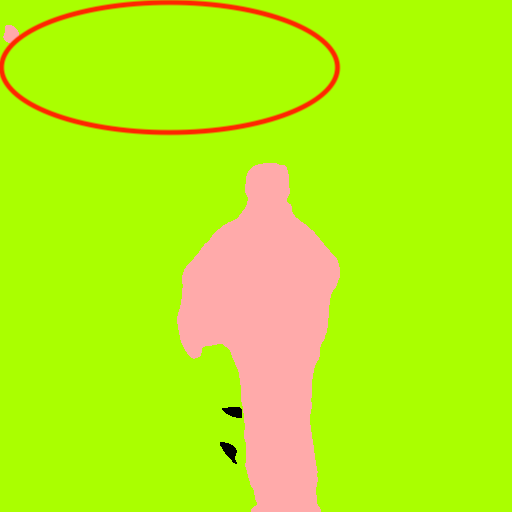}\includegraphics[width=0.2125\linewidth]{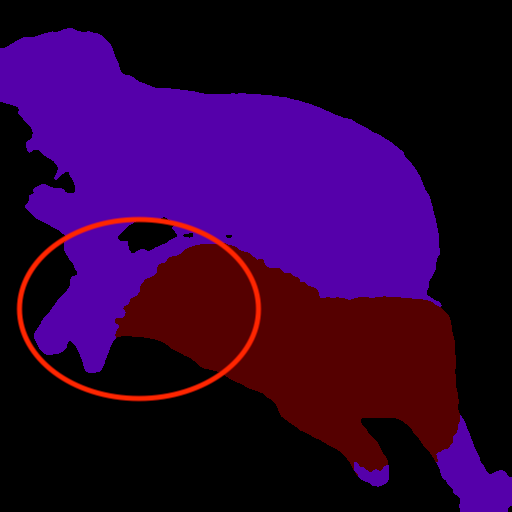}\includegraphics[width=0.2125\linewidth]{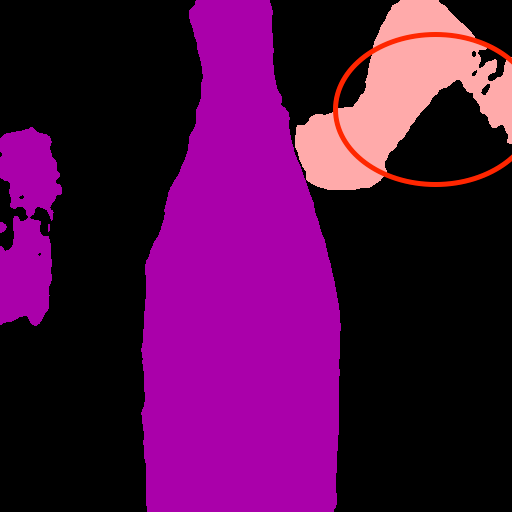}\includegraphics[width=0.2125\linewidth]{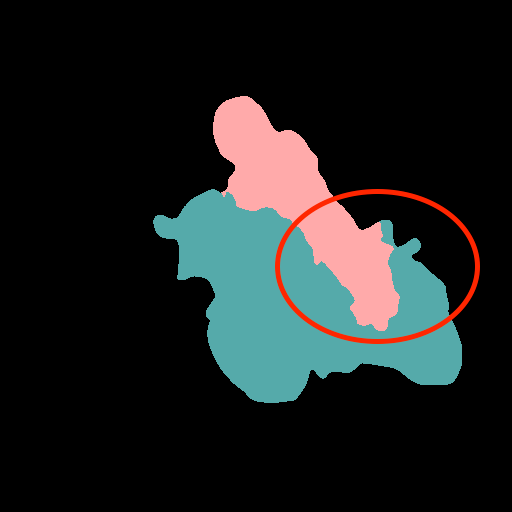}\\
\textbf{LPS on Shifted Input}\\
\includegraphics[width=0.2125\linewidth]{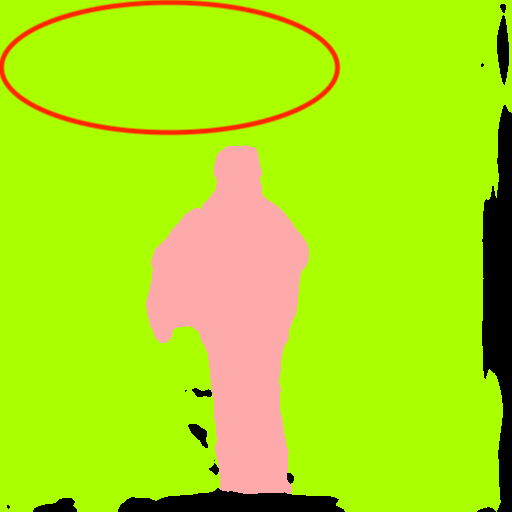}\includegraphics[width=0.2125\linewidth]{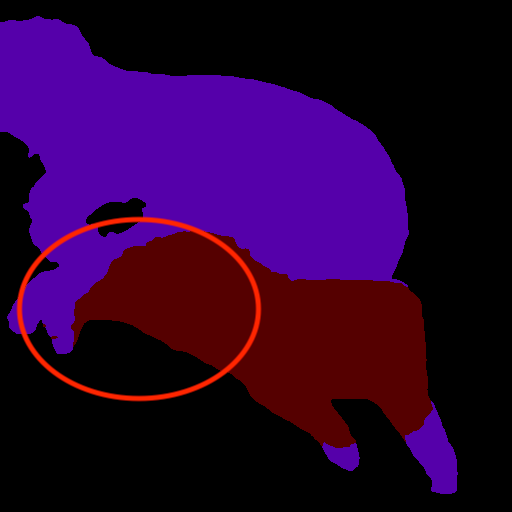}\includegraphics[width=0.2125\linewidth]{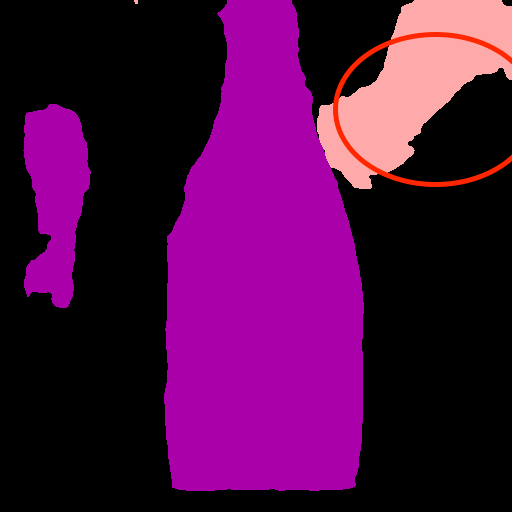}\includegraphics[width=0.2125\linewidth]{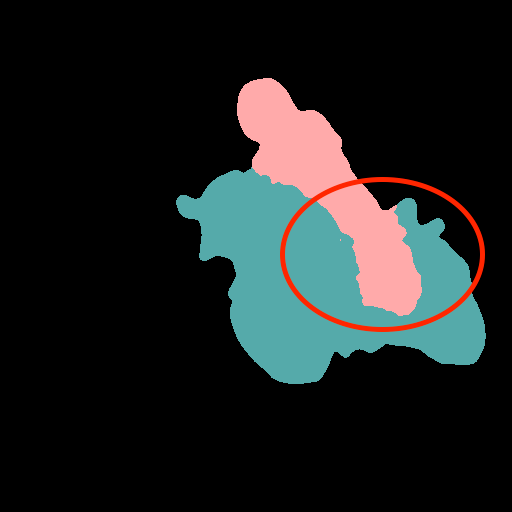}\\
\vspace{-0.2cm}
\captionof{figure}{Qualitative comparison on PASCAL VOC ResNet-101 with DeepLabV3 architecture. Regions where our proposed network showed significant improvements under linear shifts are highlighted with a red circle.}
\label{fig:semanctic_quan_11}
\end{minipage}

\begin{minipage}{1\textwidth}
\centering
\textbf{Input}\\
\includegraphics[width=0.2125\linewidth]{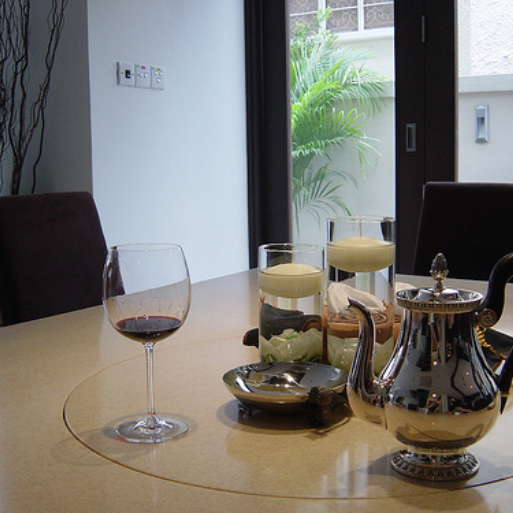}\includegraphics[width=0.2125\linewidth]{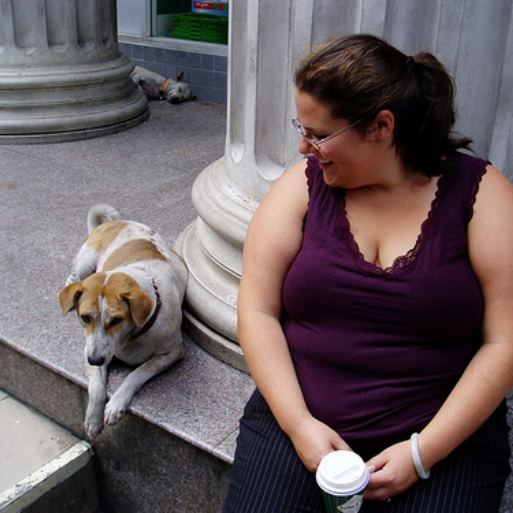}\includegraphics[width=0.2125\linewidth]{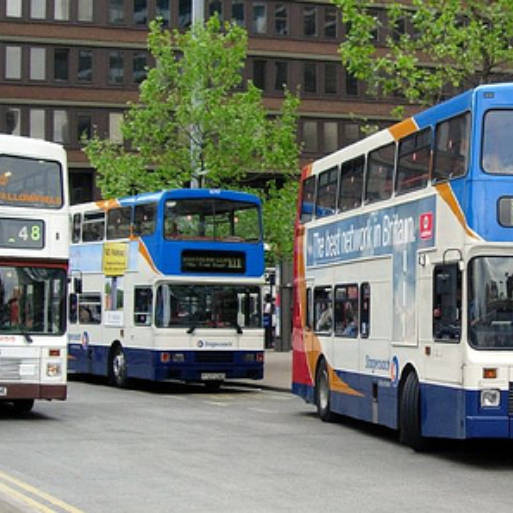}\includegraphics[width=0.2125\linewidth]{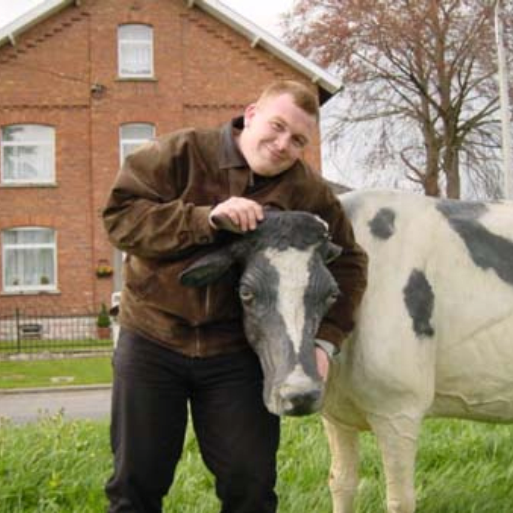}\\
\textbf{DDAC on Input}\\
\includegraphics[width=0.2125\linewidth]{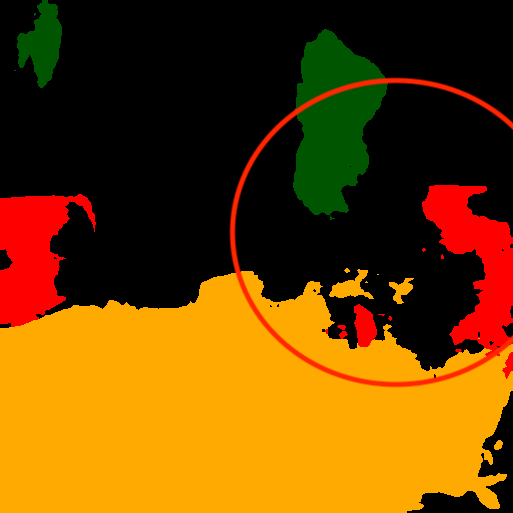}\includegraphics[width=0.2125\linewidth]{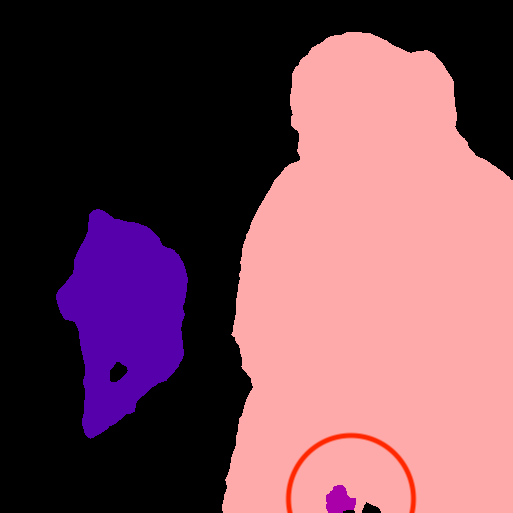}\includegraphics[width=0.2125\linewidth]{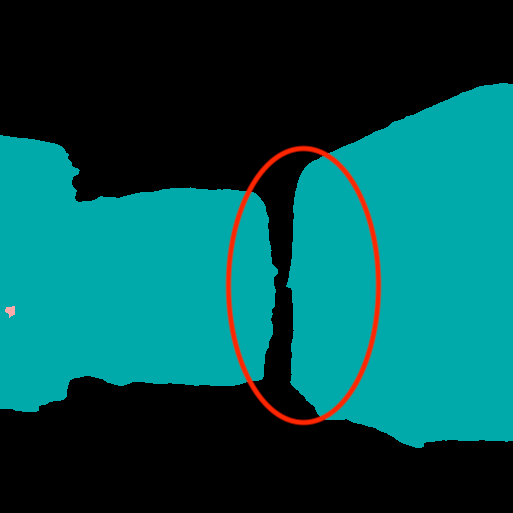}\includegraphics[width=0.2125\linewidth]{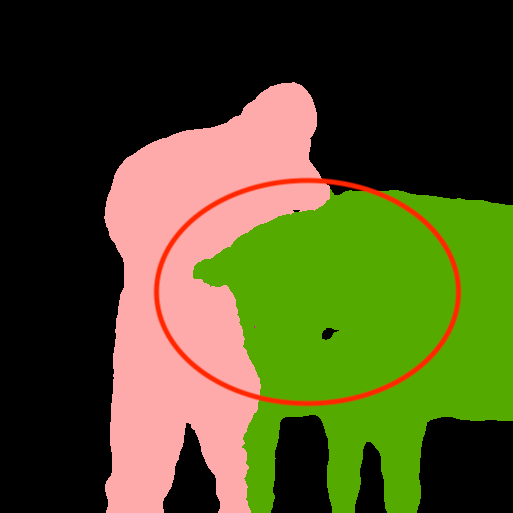}\\
\textbf{DDAC on Shifted Input}\\
\includegraphics[width=0.2125\linewidth]{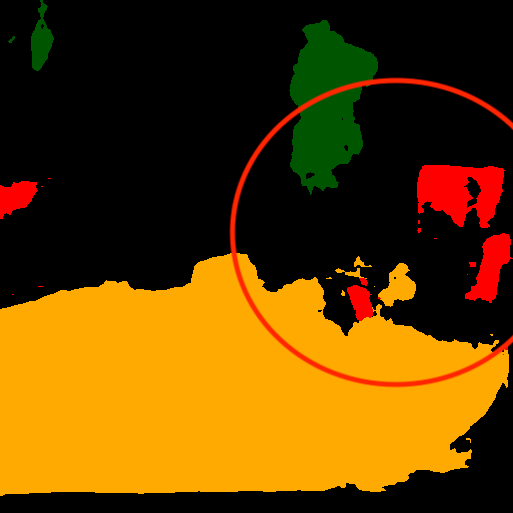}\includegraphics[width=0.2125\linewidth]{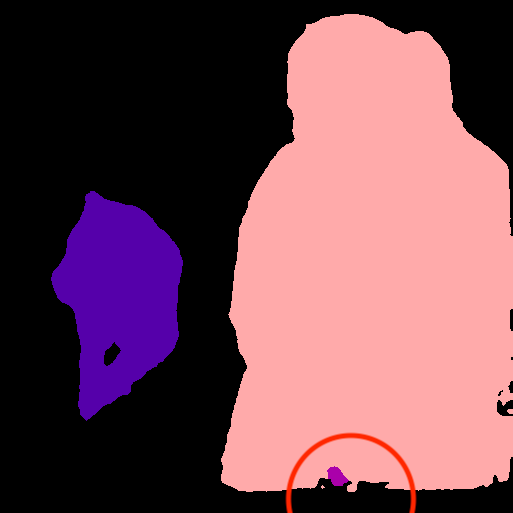}\includegraphics[width=0.2125\linewidth]{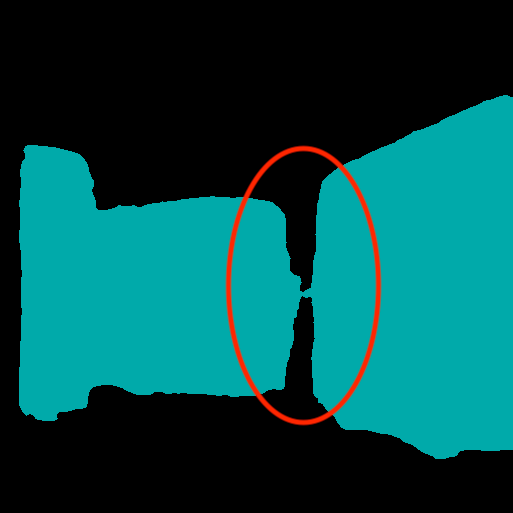}\includegraphics[width=0.2125\linewidth]{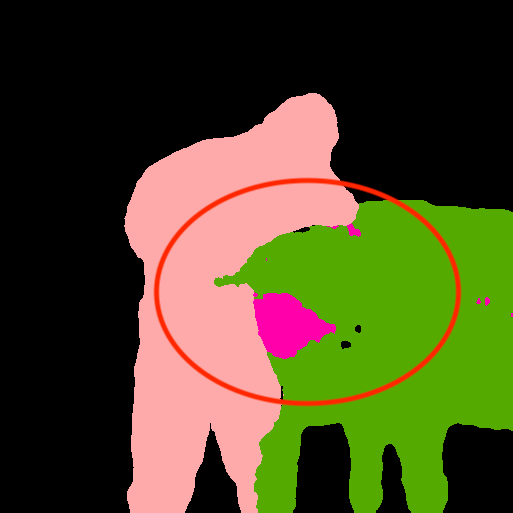}\\
\textbf{LPS on Input}\\
\includegraphics[width=0.2125\linewidth]{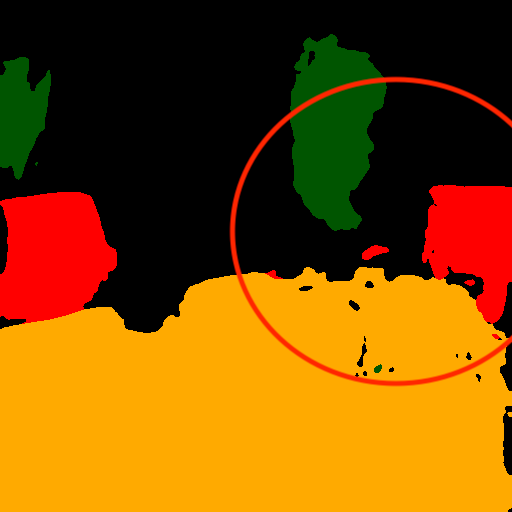}\includegraphics[width=0.2125\linewidth]{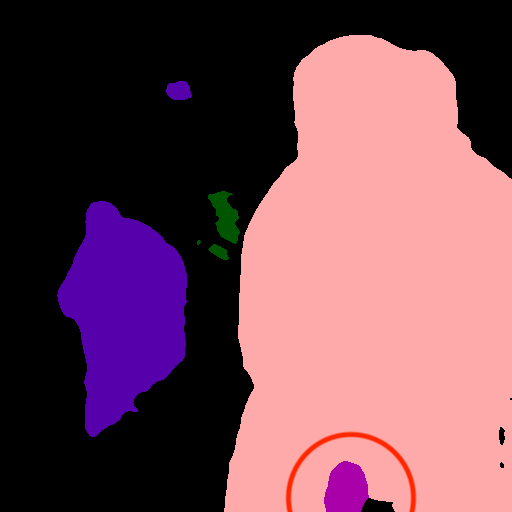}\includegraphics[width=0.2125\linewidth]{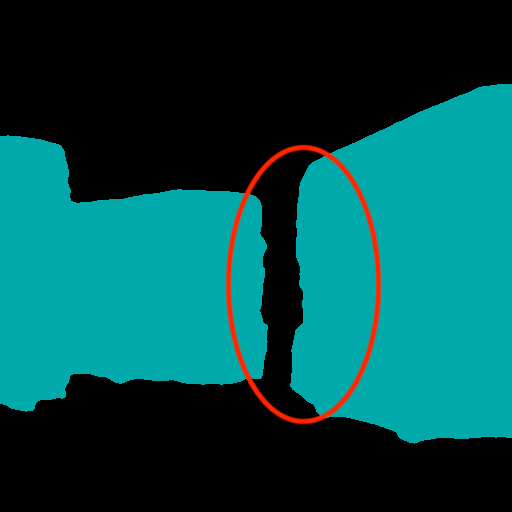}\includegraphics[width=0.2125\linewidth]{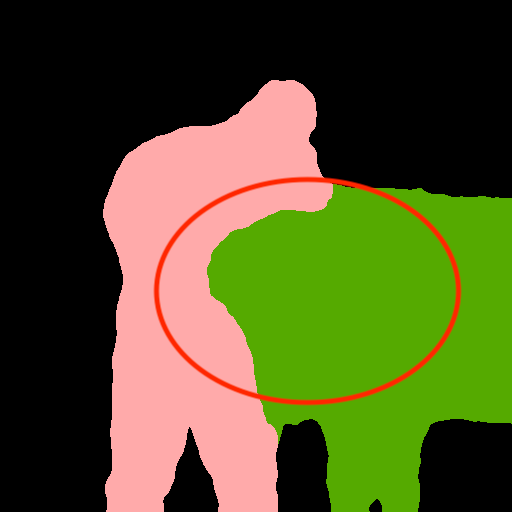}\\
\textbf{LPS on Shifted Input}\\
\includegraphics[width=0.2125\linewidth]{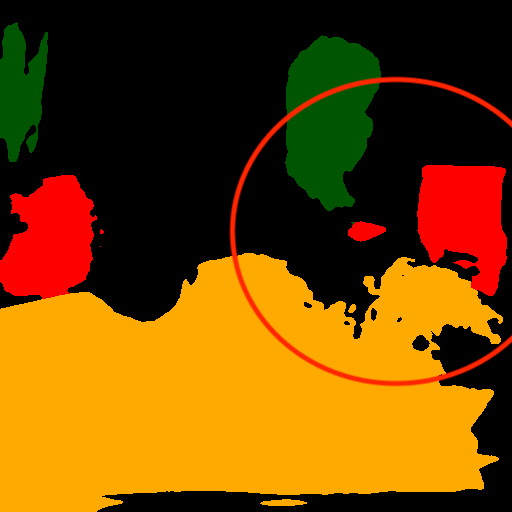}\includegraphics[width=0.2125\linewidth]{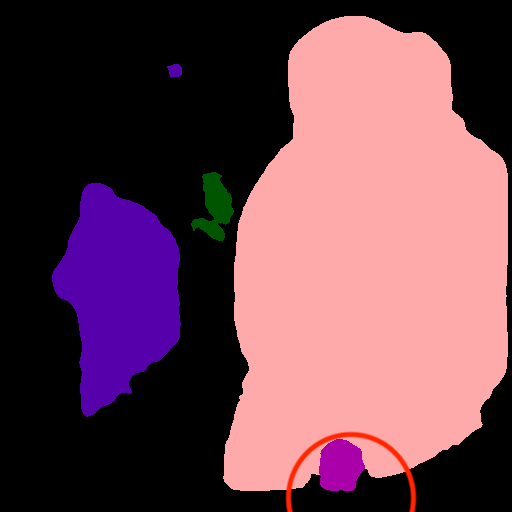}\includegraphics[width=0.2125\linewidth]{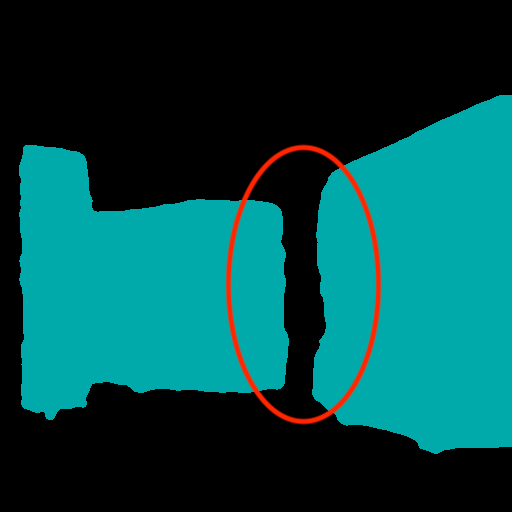}\includegraphics[width=0.2125\linewidth]{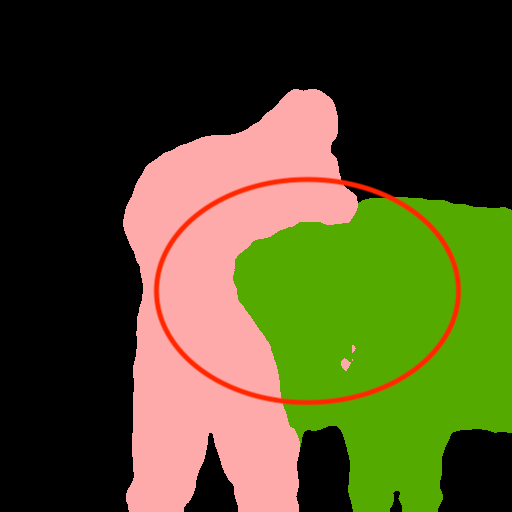}\\
\vspace{-0.2cm}
\captionof{figure}{Qualitative comparison on PASCAL VOC ResNet-101 with DeepLabV3 architecture. Regions where our proposed network showed significant improvements under linear shifts are highlighted with a red circle.}
\label{fig:semanctic_quan_12}
\end{minipage}

\section{Additional Implementation Details}\label{supp_sec:impl}
\subsection{LPS Implementation Overview}
To effectively conduct experiments, we utilize the research framework PytorchLighting~\cite{lighting} to avoid boilerplate and code redundancies. For documenting our experiments, we use Neptune~\cite{Neptune_team_neptune_ai_2019} to log experimental details and outputs to ensure reproducibility. We implement the described LPS layers in Pytorch~\cite{pytorch}. For these layers, we have written unit tests and end-to-end tests verifying the shift-invariant/equivariant properties numerically.
Overall, the attached code-base is organized as follows:
\begin{verbatim}
learn_poly_sampling
├── demo  # Ipython notebook illustrating layer usage.
├── learn_poly_sampling
│   ├── callbacks
│   ├── clargs
│   ├── configs
│   ├── data
│   ├── eval.py
│   ├── eval_segmentation.py
│   ├── layers    # LPS layers' implementation
│   ├── Makefile  # Runs tests
│   ├── models    # Classifier and DeepLabV3+ implementation
│   ├── README.md
│   ├── requirements.txt
│   ├── tests    # Test cases for the models
│   ├── train.py
│   ├── train_segmentation.py
│   └── utils
├── Makefile
└── README.md
\end{verbatim}
Please refer to ${\tt README.md}$ for installation and experimentation instructions.

\subsection{LPD Usage Illustration}
We illustrate how to incorporate the learnable polyphase downsampling (LPD) layer  into a simple classifier. The network architecture consists of a single convolution layer, followed by LPD, a global pooling and finally a fully connected layer. We numerically verify that this model is shift-invariant. As our downsampling layer is implemented as a ${\tt nn.Module}$, it can be easily incorporated into any existing deep-net implemented in Pytorch.
\begin{minted}[bgcolor=mygr]{python}
# Define Model
class SimpleClassifier(nn.Module):
    def __init__(self, num_classes=3,padding_mode='circular'):
        # Conv. Layer
        super().__init__()
        self.conv1 = nn.Conv2d(3, 32, 3, padding=1,
                               padding_mode=padding_mode)
        # Learnable Polyphase Downsampling Layer
        self.lpd = set_pool(partial(
            PolyphaseInvariantDown2D,
            component_selection=LPS,
            get_logits=get_logits_model('LPSLogitLayers'),
            pass_extras=False
            ),p_ch=32,h_ch=32)
        # Global Pooling + Classifier
        self.avgpool=nn.AdaptiveAvgPool2d((1,1))
        self.fc=nn.Linear(32, num_classes)
    def forward(self,x):
        x = self.conv1(x)
        x = self.lpd(x)  # Just like any layer.
        x = torch.flatten(self.avgpool(x),1)
        return self.fc(x)
        
# Construct Model
torch.manual_seed(0)
model = SimpleClassifier().cuda().eval().double()
# Load Image
img = torch.from_numpy(np.array(Image.open('butterfly.png'))).permute(2,0,1)
img = img.unsqueeze(0).cuda().double()
# Check is circular shift invariant
y_orig = model(img).detach().cpu()
img_roll = torch.roll(img,shifts=(1, 1), dims=(-1, -2))
y_roll = model(img_roll).detach().cpu()
print("y_orig : %
print("y_roll : %
assert(torch.allclose(y_orig,y_roll)) # Check shift invariant
print("Norm(y_orig-y_roll): %
\end{minted}
{\noindent \bf Out:}\vspace{-0.1cm}
\begin{mdframed}[backgroundcolor=myoo,  linecolor=white]
\begin{lstlisting}
y_orig : tensor([[-22.0681, -36.2678,  20.5928]],
dtype=torch.float64)
y_roll : tensor([[-22.0681, -36.2678,  20.5928]],
dtype=torch.float64)
Norm(y_orig-y_roll): 0.000000e+00
\end{lstlisting}
\end{mdframed}

\subsection{LPU Usage Illustration}
We now illustrate how to incorporate the learnable polyphase upsampling (LPU) layer  into a simple encoder-decoder architecture. The network archticture consists of a convolution layer, followed by LPD, another convolution layer, followed by LPU. We numerically verify that this architecture is circular shift-equivariant.
\vspace{-0.1cm}
\begin{minted}[bgcolor=mygr]{python}
class SimpleUNet(nn.Module):
    def __init__(self, num_classes=3,padding_mode='circular'):
        # Conv. Layer
        super().__init__()
        self.conv1 = nn.Conv2d(3, 32, 3, padding=1,
                               padding_mode=padding_mode)
        # Learnable Polyphase Downsampling Layer
        self.lpd = set_pool(partial(
            PolyphaseInvariantDown2D,
            component_selection=LPS,
            get_logits=get_logits_model('LPSLogitLayers'),
            pass_extras=False
            ),p_ch=32,h_ch=32)
        # Conv. Layer
        self.conv2 = nn.Conv2d(32, 32, 3, padding=1,
                               padding_mode=padding_mode)
        # Learnable Polyphase Upsampling Layer
        antialias_layer = get_antialias(antialias_mode='LowPassFilter',
                                        antialias_size=3,
                                        antialias_padding='same',
                                        antialias_padding_mode='circular',
                                        antialias_group=1)
        self.lpu = set_unpool(partial(
            PolyphaseInvariantUp2D,
            component_selection=LPS_u,
            antialias_layer=antialias_layer), p_ch=32)
    def forward(self,x):
        x = self.conv1(x)
        x, prob = self.lpd(x,ret_prob=True)  # Just like any layer.
        x = self.conv2(x)
        x = self.lpu(x,prob=prob) # Just like any layer.
        return x
# Construct Model
torch.manual_seed(0)
model = SimpleUNet().cuda().eval().double()
# Load Image
img = torch.from_numpy(np.array(Image.open('butterfly.png'))).permute(2,0,1)
img = img.unsqueeze(0).cuda().double()
# Check is circular shift equivariant
y_orig = model(img).detach().cpu()
img_roll = torch.roll(img,shifts=(1, 1), dims=(-1, -2))
y_roll = model(img_roll).detach().cpu()
# Roll back to check equality
y_roll_s = torch.roll(y_roll, shifts=(-1,-1), dims=(-1, -2))
print("Norm(y_orig-y_roll_s): %
assert torch.allclose(y_orig, y_roll_s)
\end{minted}
{\noindent \bf Out:}\vspace{-0.1cm}
\begin{mdframed}[backgroundcolor=myoo,  linecolor=white]
\begin{lstlisting}
Norm(y_orig-y_roll_s): 0.000000e+00
\end{lstlisting}
\end{mdframed}

\section{Additional Experimental Details}\label{supp_sec:exp}
\subsection{LPS Architecture}
As described in \secref{subsec:lps_design}, LPS selects optimal polyphase components via a mapping $f_{\theta}$ that is shift-permutation equivariant. Given a feature map $\vx\in\mathbb{R}^{C\times N_{1}\times N_{2}}$, let its polyphase decomposition of order $2$ be denoted as $\{\vx_{k}\}_{k=0}^{3}, \vx_{k}\in \mathbb{R}^{C\times \floor{N_{1}/2}\times \floor{N_{2}/2}}$. $f_{\theta}$ is then parameterized as a two-layer CNN followed by global average pooling. See~\tabref{tab:supp_lps} and \tabref{tab:supp_logits} for the LPD layer (LPD) and the logits model architecture details.

\begin{table*}[ht]
\begin{center}
\caption{\label{tab:supp_lps} Learnable polyphase downsampling (LPD) model. After computing the polyphase components and their logits, the component selection step keeps the phase with the largest logit value.}
\resizebox{1\textwidth}{!}{
\begin{tabular}{c|c|c|c|c|c|c|c|c|c}
\specialrule{.15em}{.05em}{.05em} 
& \makecell{Layer}
& \makecell{Kernel\\ Size}
& \makecell{Bias}
& \makecell{Stride}
& \makecell{Pad}
& \makecell{Input\\ Size}
& \makecell{Output\\ Size}
& \makecell{Input\\ Channels}
& \makecell{Output\\ Channels} \\
\midrule
1 & Polyphase decomposition & $-$ & $-$ & $2$ & $-$ & $N_{1}\times N_{2}$ & ${4\times \floor{N_{1}/2}\times \floor{N_{2}/2}}$ & $C$ & $C$\\
2 & Logits model & $-$ & $-$ & $-$ & $-$ & $4\times \floor{N_{1}/2}\times \floor{N_{2}/2}$ & ${4}$ & $C$ & $1$ \\
3 & Component selection & $-$ & $-$ & $-$ & $-$ & ${4\times \floor{N_{1}/2}\times \floor{N_{2}/2}}$, ${4}$ & $\floor{N_{1}/2}\times \floor{N_{2}/2}$ & $C$ & $C$\\
\specialrule{.15em}{.05em}{.05em}
\end{tabular}}
\end{center}
\vspace{-0.5\baselineskip}
\end{table*}

\begin{table*}[ht]
\begin{center}
\caption{\label{tab:supp_logits} Polyphase logits model for a single polyphase component $f_{\theta}: \mathbb{R}^{C\times \floor{N_{1}/2} \times \floor{N_{2}/2}}\mapsto \mathbb{R}$.}
\resizebox{1\textwidth}{!}{
\begin{tabular}{c|c|c|c|c|c|c|c|c|c}
\specialrule{.15em}{.05em}{.05em} 
& \makecell{Layer}
& \makecell{Kernel\\ Size}
& \makecell{Bias}
& \makecell{Stride}
& \makecell{Pad}
& \makecell{Input\\ Size}
& \makecell{Output\\ Size}
& \makecell{Input\\ Channels}
& \makecell{Output\\ Channels} \\
\midrule
1 & Conv2d + ReLU & $3\times 3$ & \cmark & $1$ & $1$ & $\floor{N_{1}/2}\times \floor{N_{2}/2}$ & $\floor{N_{1}/2}\times \floor{N_{2}/2}$ & $C$ & $C_{\text{hid}}$ \\
\midrule
2 & Conv2d & $3\times 3$ & \cmark & $1$ & $1$ & $\floor{N_{1}/2}\times \floor{N_{2}/2}$ & $\floor{N_{1}/2}\times \floor{N_{2}/2}$ & $C_{\text{hid}}$ & $C_{\text{hid}}$ \\
\midrule
3a & Flatten & $-$ & $-$ & $-$ & $-$ & $\floor{N_{1}/2}\times \floor{N_{2}/2}$ & $C_{\text{hid}}\floor{N_{1}/2} \floor{N_{2}/2}$ & $C_{\text{hid}}$ & $1$ \\
3b & Global average pooling & $-$ & $-$ & $-$ & $-$ & $C_{\text{hid}}\floor{N_{1}/2} \floor{N_{2}/2}$ & $1$ & $1$ & $1$ \\
\specialrule{.15em}{.05em}{.05em}
\end{tabular}}
\end{center}
\vspace{-0.5\baselineskip}
\end{table*}

\paragraph{Hidden Layer.} In practice, we have the freedom of choosing the number of channels in the hidden layer, denoted as $C_{\text{hid}}$. For our classification results, $C_{\text{hid}}$ is equivalent to the number of channels of the input tensor. The only exception of this rule corresponds to the LPS ResNet-101 (adaptive antialias filter) case reported in~\tabref{tab:imagenet_standard}. A top-1 classification accuracy of $78.8\%$ plus a standard shift consistency of $92.4\%$ is achieved by reducing the number of hidden channels \wrt the input at each pooling layer. Please see~\tabref{tab:supp_channels} for the number of input and hidden channels for each LPS layer used in our ResNet-101 experiments.
\begin{table*}[ht]
\begin{center}
\caption{\label{tab:supp_channels} ResNet-101: Number of channels at each LPS-D layer.}
\resizebox{0.9\textwidth}{!}{
\begin{tabular}{c|c|c|c|c|c}
\specialrule{.15em}{.05em}{.05em} 
\makecell{Layer}
& \makecell{Conv 1}
& \makecell{Maxpool}
& \makecell{Layer 2\\ Downsample}
& \makecell{Layer 3\\ Downsample}
& \makecell{Layer 4\\ Downsample}
\\
\midrule
Input Channels $C$ & $64$ & $64$ & $128$ & $256$ & $1024$\\
\midrule
Hidden Channels $C_{\text{hid}}$ & $8$ & $8$ & $16$ & $32$ & $72$\\
\specialrule{.15em}{.05em}{.05em}
\end{tabular}}
\end{center}
\vspace{-0.5\baselineskip}
\end{table*}

\paragraph{Input Dimensionality.} Under a mini-batch training setting, the polyphase decomposition of an input corresponds to a five-dimensional tensor. Let this be denoted as $\mathbf{X}\in \mathbb{R}^{B\times P\times C\times \floor{\frac{N_{1}}{2}}\times \floor{\frac{N_{2}}{2}}}$, where $B$ corresponds to the total number of feature maps in the batch and $P=4$ to the number of polyphase components (assuming a downscaling factor of $2$). To efficiently obtain a logit for each component, independently of its relative position in the tensor, we reshape it by combining the batch and polyphase component dimensions.

This alternative representation corresponds to $\hat{\mathbf{X}}\in \mathbb{R}^{4B\times C\times \floor{\frac{N_{1}}{2}}\times \floor{\frac{N_{2}}{2}}}$, and allows for each polyphase component to be processed independently of the rest. In practice, our CNN-based logits model receives $\hat{\mathbf{X}}$ as input and generates a set of $4B$ logits, one for each polyphase component, in a single forward pass.

{\bf\noindent Overall Architecture.}
\tabref{tab:supp_resnet50} and \tabref{tab:supp_reslayer} provide a general description of the ResNet-based architecture that incorporates LPD as pooling layer and its custom residual layer, respectively. In contrast to the original ResNet model, each pooling or downsampling step is replaced by our learn-based layer. For illustration purposes, we focus on the ResNet-50 model. 
\begin{table*}[t]
\begin{center}
\caption{\label{tab:supp_resnet50}  LPS-based ResNet-50 architecture for ImageNet.}
\resizebox{1\textwidth}{!}{
\begin{tabular}{c|c|c|c|c|c|c|c|c|c|c}
\specialrule{.15em}{.05em}{.05em} 
& \makecell{Layer}
& \makecell{Kernel\\ Size}
& \makecell{Bias}
& \makecell{Stride}
& \makecell{Pad}
& \makecell{Input\\ Size}
& \makecell{Output\\ Size}
& \makecell{Hidden\\ Channels}
& \makecell{Input\\ Channels}
& \makecell{Output\\ Channels} \\
\midrule
1a & Conv2d + BN + ReLU & $7\times 7$ & \xmark & $1$ & $3$ & $224\times 224$ & $224\times 224$ & $-$ & $3$ & $64$ \\
1b & LPD & $-$ & $-$ & $2$ & $-$ & $224\times 224$ & $112\times 112$ & $-$ & $64$ & $64$ \\
\midrule
2a & Zero pad & $-$ & $-$ & $-$ & $-$ & $112\times 112$ & $113\times 113$ & $-$ & $64$ & $64$ \\
2b & Max filter & $2\times 2$ & $-$ & $1$ & $0$ & $113\times 113$ & $112\times 112$ & $-$ & $64$ & $64$ \\
2c & LPD & $-$ & $-$ & $-$ & $-$ & $112\times 112$ & $56\times 56$ & $64$ & $64$ & $64$ \\
\midrule
\multicolumn{11}{c}{Block $1$}\\
\midrule
3a & Res. layer & $-$ & $-$ & $1$ & $-$ & $56\times 56$ & $56\times 56$ & $64$ & $64$ & $256$ \\
3b & Res. layer & $-$ & $-$ & $1$ & $-$ & $56\times 56$ & $56\times 56$ & $64$ & $256$ & $256$ \\
3c & Res. layer & $-$ & $-$ & $1$ & $-$ & $56\times 56$ & $56\times 56$ & $64$ & $256$ & $256$ \\
\midrule
\multicolumn{11}{c}{Block $2$}\\
\midrule
4a & Res. layer (LPD) & $-$ & $-$ & $2$ & $-$ & $56\times 56$ & $28\times 28$ & $128$ & $256$ & $512$ \\
4b & Res. layer & $-$ & $-$ & $1$ & $-$ & $28\times 28$ & $28\times 28$ & $128$ & $512$ & $512$ \\
4c & Res. layer & $-$ & $-$ & $1$ & $-$ & $28\times 28$ & $28\times 28$ & $128$ & $512$ & $512$ \\
4d & Res. layer & $-$ & $-$ & $1$ & $-$ & $28\times 28$ & $28\times 28$ & $128$ & $512$ & $512$ \\
\midrule
\multicolumn{11}{c}{Block $3$}\\
\midrule
5a & Res. layer (LPD) & $-$ & $-$ & $2$ & $-$ & $28\times 28$ & $14\times 14$ & $256$ & $512$ & $1024$ \\
5b & Res. layer & $-$ & $-$ & $1$ & $-$ & $14\times 14$ & $14\times 14$ & $256$ & $1024$ & $1024$ \\
5c & Res. layer & $-$ & $-$ & $1$ & $-$ & $14\times 14$ & $14\times 14$ & $256$ & $1024$ & $1024$ \\
5d & Res. layer & $-$ & $-$ & $1$ & $-$ & $14\times 14$ & $14\times 14$ & $256$ & $1024$ & $1024$ \\
5e & Res. layer & $-$ & $-$ & $1$ & $-$ & $14\times 14$ & $14\times 14$ & $256$ & $1024$ & $1024$ \\
5f & Res. layer & $-$ & $-$ & $1$ & $-$ & $14\times 14$ & $14\times 14$ & $256$ & $1024$ & $1024$ \\
\midrule
\multicolumn{11}{c}{Block $4$}\\
\midrule
6a & Res. layer (LPD) & $-$ & $-$ & $2$ & $-$ & $14\times 14$ & $7\times 7$ & $512$ & $1024$ & $2048$ \\
6b & Res. layer & $-$ & $-$ & $1$ & $-$ & $7\times 7$ & $7\times 7$ & $512$ & $2048$ & $2048$ \\
6c & Res. layer & $-$ & $-$ & $1$ & $-$ & $7\times 7$ & $7\times 7$ & $512$ & $2048$ & $2048$ \\
\midrule
7 & Global average pool & $-$ & $-$ & $-$ & $-$ & $7\times 7$ & $1\times 1$ & $-$ & $2048$ & $2048$ \\
8 & Flatten & $-$ & $-$ & $-$ & $-$ & $1\times 1$ & $2048$ & $-$ & $2048$ & $1$ \\
9 & Fully connected & $-$ & \cmark & $-$ & $-$ & $2048$ & $1000$ & $-$ & $1$ & $1$ \\
\specialrule{.15em}{.05em}{.05em}
\end{tabular}}
\end{center}
\end{table*}

\begin{table*}[t]
\begin{center}
\caption{\label{tab:supp_reslayer}  Example of an LPS-based residual layer. Architecture corresponds to the first residual layer of ResNet-50 block $2$ (4a in \tabref{tab:supp_resnet50}). The main and shortcut branches receive the input feature map of dimensions $N_{1}\times N_{2}$. The LPD layer in the shortcut branch also receives the logits precomputed on the main branch of dimensions $4$ to consistently select the same component.
}
\resizebox{1\textwidth}{!}{
\begin{tabular}{c|c|c|c|c|c|c|c|c|c}
\specialrule{.15em}{.05em}{.05em} 
& \makecell{Layer}
& \makecell{Kernel\\ Size}
& \makecell{Bias}
& \makecell{Stride}
& \makecell{Pad}
& \makecell{Input\\ Size}
& \makecell{Output\\ Size}
& \makecell{Input\\ Channels}
& \makecell{Output\\ Channels} \\
\midrule
\multicolumn{10}{c}{Main branch}\\
\midrule
1a & Conv2d + BN + ReLU & $1\times 1$ & \xmark & $1$ & $0$ & ${N_{1}\times N_{2}}$ & $N_{1}\times N_{2}$ & $256$ & $128$ \\
1b & Conv2d + BN + ReLU & $3\times 3$ & \xmark & $1$ & $1$ & $N_{1}\times N_{2}$ & $N_{1}\times N_{2}$ & $128$ & $128$ \\
1c & LPD & $-$ & $-$ & $2$ & $-$ & $N_{1}\times N_{2}$ & $\floor{N_{1}/2}\times \floor{N_{2}/2}$, ${4}$ & $128$ & $128$\\
1d & Conv2d + BN & $1\times 1$ & \xmark & $1$ & $0$ & $\floor{N_{1}/2}\times \floor{N_{2}/2}$ & ${\floor{N_{1}/2}\times \floor{N_{2}/2}}$ & $128$ & $512$ \\
\midrule
\multicolumn{10}{c}{Shortcut branch}\\
\midrule
2a & LPD (pre-computed) & $-$ & $-$ & $2$ & $-$ & ${N_{1}\times N_{2}}$, ${4}$ & $\floor{N_{1}/2}\times \floor{N_{2}/2}$ & $256$ & $256$\\
2b & Conv2d + BN & $1\times 1$ & \xmark & $1$ & $0$ & $\floor{N_{1}/2}\times \floor{N_{2}/2}$ & ${\floor{N_{1}/2}\times \floor{N_{2}/2}}$ & $256$ & $512$ \\
\midrule
3 & Sum + ReLU & $-$ & $-$ & $-$ & $-$ & ${\floor{N_{1}/2}\times \floor{N_{2}/2}}$, ${\floor{N_{1}/2}\times \floor{N_{2}/2}}$ & $\floor{N_{1}/2}\times \floor{N_{2}/2}$ & $512$ & $512$ \\
\specialrule{.15em}{.05em}{.05em}
\end{tabular}}
\end{center}
\vspace{-0.2cm}
\end{table*}

\subsection{Baseline Implementations}
\subsubsection{Image Classification}
In this section, we provide additional implementation details and differences from the three main classification baselines, Lowpass Filtering (LPF) \cite{zhang2019making}, Adaptive Polyphase Sampling (APS) \cite{chaman2021truly} and Adaptive Lowpass Filtering (DDAC) \cite{zou2020delving}.

\paragraph{Lowpass Filtering (LPF).}  LPF classification accuracy and shift consistency values included in \tabref{tab:cifar10}, \tabref{tab:imagenet_circular} and \tabref{tab:imagenet_standard} correspond to those reported by LPF and APS manuscripts. Experimental results for standard shift consistency were taken from the LPF official repository, while experimental results analyzing circular shift consistency correspond to those reported by APS. It is important to note that, while LPF training setup for standard shift consistency uses \textit{rescaled random cropping} as part of its preprocessing, our experiments on circular shift consistency follow APS settings and discard it. Please refer to \secref{sec:supp_hyperparams} for details regarding data preprocessing.

\paragraph{Adaptive Polyphase Sampling (APS).} As with LPF, APS's accuracy and consistency values via no antialiasing or lowpass filtering included in~\secref{sec:exp} are obtained from their official manuscript.  APS results via adaptive filtering, denoted in~\tabref{tab:cifar10} and~\tabref{tab:imagenet_circular} were obtained by replacing our learnable polyphase selection criteria by APS $\ell_{2}$ energy-based selection and incorporated the adaptive filtering to it.

\paragraph{Adaptive Lowpass Filtering (DDAC).} We compare the accuracy and shift consistency of our proposed pooling method against that of DDAC under both circular and standard shifts. For the circular shift case, reported for CIFAR-10 and ImageNet in \tabref{tab:cifar10} and \tabref{tab:imagenet_circular}, respectively, we replace the LPS selection criteria by keeping always the even polyphase components ($k^{\star}=0$), followed by a learnable low-pass filter with the exact same specifications as the one provided in the official DDAC code base. For consistency purposes, DDAC experiments analyzing circular shift consistency follow the same data preprocessing used in APS experiments. More precisely, no rescaled random cropping is applied for data augmentation.

For the standard shift case, we compare against the best performing DDAC model, ResNet-101, without any changes. \tabref{tab:imagenet_standard} includes its reported top-1 classification accuracy and shift consistency. For the sake of clarity, we also include the results obtained by training the model from scratch using their official code base and hyperparameters.

\subsubsection{Semantic Segmentation}
In the paper, we directly compared to the reported values in DDAC~\cite{zou2020delving}. Unfortunately, the authors did not released their evaluation code and did not provide a clear description of the metric. We were not able to exactly reproduce their reported mASSC. For a fair comparison, we evaluate DDAC's released checkpoint using our mASSC implementation. We will make our implementation publicly available.

\subsection{Comparison against Classifiers with More Trainable Parameters.}
\label{sec:supp_comparison}
To show the performance improvement attained by our LPD approach is not simply an effect of introducing more trainable parameters, we compared our ResNet-101 + LPD model ($44,751,034$ parameters) against the larger ResNet-152 model ($60,192,808$ parameters). Both models were trained on ImageNet using the same augmentation and optimizer configuration. Under such settings, ResNet-152 achieves 78.3\% top-1 classification accuracy and 90.9\% shift consistency, while our ResNet-101 + LPD model obtains 78.8\% top-1 accuracy and 92.4\% shift-consistency. Despite having 25\% less trainable parameters, our model attains 0.5\% higher accuracy and 1.5\% higher shift-consistency.

\subsection{Hyperparameters and Tuning Procedure}\label{sec:supp_hyperparams}
\subsubsection{Image Classification}
\paragraph{Learn Rate and Optimization Parameters.} Following the standard ImageNet setup, the initial learn rate value corresponds to $0.1$ and follows a multi-step schedule, decaying every $30$ epochs by a factor of $0.1$. Our models are trained via stochastic gradient descent with $0.9$ momentum. A weight decay of factor $10^{-4}$ is imposed to all model trainable weights except those of the LPS layers. Empirically, this has shown a substantial consistency improvement, avoiding cases where polyphase logits have very similar values and other numerical precision issues. Additionally, for our experiments on ResNet-101 and ResNet-50 with adaptive antialiasing filters, following DDAC settings, a learning rate warmup of five epochs is applied.

\paragraph{Data Preprocessing and Split.} \tabref{tab:supp_preproc_cifar10} describes the data preprocessing used in our CIFAR-10 experiments.
No shifts or resizing augmentations are applied to highlight the fact that perfect circular shift invariance/equivariance is achieved by design and not induced during training.

\begin{table}[ht]
\begin{center}
\caption{\label{tab:supp_preproc_cifar10} CIFAR-10 data preprocessing.}
\resizebox{0.7\textwidth}{!}{
\begin{tabular}{c|c|c}
\specialrule{.15em}{.05em}{.05em} 
\makecell{Split} & Train Set & Test Set\\
\midrule
\makecell{Preprocessing} & \makecell{(i) Random horizontal flipping\\ (ii) Normalization}
& \makecell{(i) Normalization}\\
\specialrule{.15em}{.05em}{.05em}
\end{tabular}}
\end{center}
\vspace{-0.2cm}
\end{table}

For ImageNet experiments evaluating circular shift consistency, we follow APS's preprocessing settings.~\tabref{tab:supp_preproc_imagenet_circ} describes its data preprocessing. For ImageNet experiments evaluating standard shift consistency, we follow DDAC settings. \tabref{tab:supp_preproc_imagenet_std} describes its data preprocessing.
\begin{table}[ht]
\begin{center}
\caption{\label{tab:supp_preproc_imagenet_circ} ImageNet data preprocessing for \textbf{circular shift consistency} evaluation.}
\resizebox{0.8\textwidth}{!}{
\begin{tabular}{c|c|c}
\specialrule{.15em}{.05em}{.05em} 
\makecell{Split} & Train Set & Test Set\\
\midrule
\makecell{Preprocessing} & \makecell{(i) Resizing to $256 \times 256$\\(ii) Center cropping to $224 \times 224$\\ (iii) Random horizontal flipping\\ (iv) Normalization}
& \makecell{(i) Resizing to $256 \times 256$\\(ii) Center cropping to $224 \times 224$\\ (iii) Normalization}\\
\specialrule{.15em}{.05em}{.05em}
\end{tabular}}
\end{center}
\vspace{-0.5\baselineskip}
\end{table}

\begin{table}[H]
\begin{center}
\caption{\label{tab:supp_preproc_imagenet_std} ImageNet data preprocessing for \textbf{standard shift consistency} evaluation.}
\resizebox{0.8\textwidth}{!}{
\begin{tabular}{c|c|c}
\specialrule{.15em}{.05em}{.05em} 
\makecell{Split} & Train Set & Test Set\\
\midrule
\makecell{Preprocessing} & \makecell{(i) Resizing to $256 \times 256$\\(ii) Resized random cropping to $224 \times 224$\\ (iii) Random horizontal flipping\\ (iv) Normalization}
& \makecell{(i) Resizing to $256 \times 256$\\(ii) Center cropping to $224 \times 224$\\ (iii) Normalization}\\
\specialrule{.15em}{.05em}{.05em}
\end{tabular}}
\end{center}
\vspace{-0.5\baselineskip}
\end{table}

\paragraph{Computational Settings.} Classification experiments on CIFAR-10 are trained on a single NVIDIA Titan V using a batch size of $256$ for $250$ epochs. Classification experiments on ImageNet are trained in distributed data parallel mode on four NVIDIA A6000 GPUs using a batch size of $64$ for $90$ epochs.

\paragraph{Polyphase Selection for Circular Shift Consistency.} For circular shift consistency on both CIFAR-10 and ImageNet, the polyphase component selection depends on the model state. During training, each LPS layer samples from a Gumbel-softmax distribution. This leads to a convex combination of polyphase components that improves the backpropagation process. Following the original Gumbel-Softmax formulation, we use an annealing factor to slowly converge to a one-hot vector along epochs. Considering $250$ and $90$ training epochs for CIFAR-10 and ImageNet experiments, respectively, the annealing factor $\tau \in \mathbb{R}_{+}$ gradually decays to improve the gradient flow during the error backpropagation step. A step decay approach is used for CIFAR-10 experiments, while a multistep linear decay is used for ImageNet experiments.

During testing, optimal polyphase components correspond to those with the largest logit values (hard-selection), which leads to a classifier with perfect circular shift consistency by design.

\paragraph{Polyphase Selection for Standard Shift Consistency.} For standard shift consistency evaluation on ImageNet, we adopt a fine-tuning procedure to balance the shift consistency and classification accuracy obtained by our models. Recall that our model guarantees perfect consistency under circular shifts, which leaves open the posibility of applying more refined training to improve the performance under standard shifts.

With this in mind, instead of switching to a hard-selection, we relax the annealing factor during the last $28$ training epochs, replacing the gumbel-softmax sampling by a standard softmax (soft-association) for both training and testing.

\tabref{tab:supp_tau} includes the details of the Gumbel-softmax annealing schedule for both circular and standard shift consistency experiments.

\begin{table*}[ht]
\begin{center}
\caption{\label{tab:supp_tau} Gumbel-softmax annealing schedule for image classification.}
\resizebox{0.8\textwidth}{!}{
\begin{tabular}{c|c|c|c}
\specialrule{.15em}{.05em}{.05em} 
\multirow{2}{*}{\makecell{Consistency\\Evaluation}} & CIFAR-10 & \multicolumn{2}{c}{ImageNet}\\
\cline{2-4}
& Circular Shift & Circular Shift & Standard Shift\\
\midrule
Method & Step Decay & Multirate Linear Decay & Multirate Linear Decay\\
\midrule
\makecell{$\tau$\\Schedule} & \makecell{Initial value: $\tau$=1\\Decay step: $10$ epochs\\Decay factor: $0.85$\\Minumum value: $\tau=0.025$} &
\makecell{Epoch $1/90$: $\tau=1$\\Epoch $62/90$: $\tau=0.5$\\Epoch $82/90$: $\tau=0.05$\\Epoch $90/90$: $\tau=0.01$} &
\makecell{Epoch $1/90$: $\tau=1$\\Epoch $62/90$: $\tau=0.5$\\Epoch $90/90$: $\tau=0.25$}\\
\specialrule{.15em}{.05em}{.05em}
\end{tabular}}
\end{center}
\vspace{-0.5\baselineskip}
\end{table*}

\subsubsection{Semantic Segmentation}
\paragraph{Unpooling Component Selection.} During training, for both standard and circular shift consistency evaluation, feature maps obtained from the backbone are unpooled (upsampled and shifted) by LPU layers. The shift applied to each feature map, intended to place them back into their original indices, depends on the logit probabilities computed by the backbone. Since these are obtained from a Gumbel-softmax distribution during training, instead of placing the upsampled feature map at a single position, LPU layers generate an upscaled representation composed by all four possible positions (assuming an upscaling factor of $2$), each weighted by its corresponding probability. In other words, LPU \textit{soft}-unpools the input feature map to all four possible positions, weights them by their probabilities and adds them together to obtain the output feature map.

For our ImageNet experiments we modify the annealing schedule used for image classification and tailor it to the segmentation model and its $125$ training epochs. First, we linearly increase the annealing factor from the last value used during the backbone training process (recall that the backbone was trained using its own Gumbel-softmax annealing schedule) to $0.5$. Then, we gradually decay it to improve the gradient flow during backpropagation.

During testing, LPU layers receive logits from the backbone. Then, feature maps are unpooled according to the polyphase component with the largest logit value (hard-selection), allowing the segmentation model to become shift-equivariant by design.

\tabref{tab:supp_seg_tau} includes the details of the Gumbel-softmax annealing schedule used in our semantic segmentation experiments for both circular and standard shift consistency evaluation.

\begin{table*}[t]
\begin{center}
\caption{\label{tab:supp_seg_tau} Gumbel-softmax annealing schedule for image segmentation on ImageNet. $^{*}$ Initial $\tau_{\text{backbone}}$ values correspond to the final values used during our ResNet-18 and ResNet-101 backbone training.}
\resizebox{0.8\textwidth}{!}{
\begin{tabular}{c|c|c}
\specialrule{.15em}{.05em}{.05em} 
\multirow{2}{*}{\makecell{Consistency\\Evaluation}} & ResNet-18 & ResNet-101\\
\cline{2-3}
& Circular Shift & Circular and Standard Shift\\
\midrule
Method & Multirate Linear Decay & Multirate Linear Decay\\
\midrule
\makecell{$\tau$\\Schedule} & \makecell{Epoch $1/125$: $\tau=\tau_{\text{backbone}}$  $^{*}$\\Epoch $60/125$: $\tau=0.5$\\Epoch $82/125$: $\tau=0.15$\\Epoch $90/125$: $\tau=0.01$} &
\makecell{Epoch $1/125$: $\tau=\tau_{\text{backbone}}$ $^{*}$\\Epoch $10/125$: $\tau=0.5$\\Epoch $80/125$: $\tau=0.3$\\Epoch $100/125$: $\tau=0.125$\\Epoch $125/125$: $\tau=0.01$}\\
\specialrule{.15em}{.05em}{.05em}
\end{tabular}}
\end{center}
\vspace{-0.5\baselineskip}
\end{table*}

\end{document}